\pdfoutput=1
\documentclass[a4paper, 11pt]{article}
\usepackage{srcltx,graphicx}
\usepackage{amsmath, amssymb, amsthm}
\usepackage{color, xcolor}
\usepackage{array}
\usepackage{lscape}
\usepackage{multirow}
\usepackage{psfrag}
\usepackage[hang]{subfigure}
\usepackage{esint}
\usepackage{bigints}
\graphicspath{{images/}}

\newtheorem{theorem}{Theorem}

{\theoremstyle{remark} }

\newtheorem{proposition}{Proposition}

\numberwithin{equation}{section}

\newtheorem{assump}[theorem]{Assumption}

\DeclareMathOperator*{\argmin}{arg\,min}

\usepackage[utf8]{inputenc} 
\usepackage[T1]{fontenc}    

\usepackage{url}            
\usepackage{booktabs}       
\usepackage{amsfonts}       
\usepackage{nicefrac}       
\usepackage{microtype}      

\begin{document}

\title{Stochastic modified equations for the asynchronous stochastic gradient descent}

\author{
Jing An\thanks{Institute for Computational and Mathematical
    Engineering, Stanford University, Stanford, CA 94305, email: {\tt
    jingan@stanford.edu}},~~
Jianfeng Lu\thanks{Department of Mathematics, 
  Department of Chemistry and Department of Physics,
  Duke University, Durham, NC 27708, email: {\tt jianfeng@math.duke.edu}},~~
Lexing Ying\thanks{Department of Mathematics and Institute for
    Computational and Mathematical Engineering, Stanford University,
    Stanford, CA 94305, email: {\tt lexing@stanford.edu}}
}
\date{}
\maketitle

\begin{abstract}
 {We propose stochastic modified equations (SMEs) for modeling the asynchronous stochastic gradient
  descent (ASGD) algorithms. The resulting SME of Langevin type extracts more information about the
  ASGD dynamics and elucidates the relationship between different types of stochastic gradient
  algorithms. We show the convergence of ASGD to the SME in the continuous time limit, as well as
  the SME's precise prediction to the trajectories of ASGD with various forcing terms. As an
  application, we propose an optimal mini-batching strategy for ASGD via solving the
  optimal control problem of the associated SME.}

\end{abstract}

\section{Introduction}

In this paper, we consider the following empirical risk minimization problem commonly encountered in
machine learning:
\begin{align}
    \min_{x\in \mathbb{R}^d} f(x):= \frac{1}{n}\sum_{i=1}^n f_i(x),
\end{align}
where $x$ represents the model parameters, $f_i(x) \equiv f(x; z_i)$ denotes the loss function of
the training sample $z_i$, and $n$ is the size of the training sample set. Since the training set
for most applications is of large size, stochastic gradient descent (SGD) is the most popular
algorithm used in practice. In the simplest scenario, SGD samples one random instance $f_i(\cdot)$
uniformly at each iteration and updates the parameter by evaluating only the gradient of the selected
$f_i(\cdot)$. The stability and convergence rate of SGD have been studied in depth, for example, see \cite{hardt2015train, needell2014stochastic}. However, the scalability of SGD is unfortunately
restricted by its inherent sequential nature. To overcome this issue and hence accelerate the
convergence, there has been a line of research devoted to asynchronous parallel SGDs. In the
distributed computation scenario, an asynchronous stochastic gradient descent (ASGD) method
parallelizes the computation on multiple processing units by (1) calculating multiple gradients
simultaneously at different processors and (2) sending the results asynchronously back to the master
for updating the model parameters \cite{agarwal2011distributed,recht2011hogwild}.

\subsection{Related Work}

There has been a vast literature on the analysis of SGD, see for example Bottou et
al. \cite{bottou2016optimization} for a comprehensive review of this subject. Some widely-used
methods include AdaGrad \cite{duchi2011adaptive}, which extends SGD by adapting step sizes for
different features, RMSProp \cite{tieleman2012lecture}, which resolves AdaGrad's rapidly diminishing
learning rates issue, and Adam \cite{kingma2014adam}, which combines the advantages of both AdaGrad
and RMSProp with a parameter learning rates adaption based on the average of the second moments of
the gradients. On the other hand, relatively few studies are devoted to ASGDs.  Most of these
studies for ASGD take an optimization perspective. Hogwild!  \cite{recht2011hogwild} assumed data
sparsity in order to run parallel SGD without locking successfully. Under various smoothness
conditions on $f$ such as $f$ being strongly convex and $f_i$'s all Lipschitz, it showed that the
convergence rate can be similar to the synchronous case. Duchi et al. \cite{duchi2013estimation}
extended this result by developing an asynchronous dual averaging algorithm that allows problems to
be non-smooth and non strongly-convex as well. Mitliagkas et al. \cite{mitliagkas2016asynchrony}
observed that a standard queuing model of asynchrony correlates to the momentum, that is, 
asynchrony produces momentum in SGD updates. There are also several methods using asynchrony either 
in parallel or in a distributed way, such as asynchronous stochastic coordinate descent algorithms
\cite{liu2015asynchronous2,liu2015asynchronous,nesterov2012efficiency,richtarik2014iteration}.

Recently, Li et al. \cite{li2015stochastic} introduced the concept of the stochastic modified
equation for SGDs (referred as {\bf SME-SGD} in this report), where in the continuous-time limit an
SGD is approximated by an appropriate (overdamped) Langevin equation. Compared to most convergence
analyses that give upper bounds for (strongly) convex objects, this new framework not only provides
more precise analyses for the leading order dynamics of SGD but also suggests adaptive
hyper-parameter strategies using optimal control theory.

\subsection{Our Contributions}
We give a novel derivation of SMEs for the ASGD algorithms by introducing auxiliary variables 
to treat an effective memory term. With the derived SME models, we are able to characterize the 
dynamics of ASGD algorithms.

In Section 2, we first derive a stochastic modified equation for the asynchronous stochastic
gradient descent, denoted shortly as {\bf SME-ASGD}, for the case where each loss
function $f_i$ is quadratic. The derivation results in a Langevin equation, which by assuming its ergodicity has a unique
invariant distribution solution with a convergence rate dominated by the temperature
factor. Meanwhile, for the momentum SGD (MSGD), a similar Langevin equation denoted as {\bf
  SME-MSGD} is derived and we show that the temperature factors for both derived SME agree. This
comparison gives a Langevin dynamics explanation of why an asynchronous method gives rise to similar
behavior as compared to the momentum-based methods \cite{mitliagkas2016asynchrony}. Then by introducing a new accumulative quantity, we derive a more general SME-ASGD for the general case in
which the gradient of the loss function can be nonlinear. We show that the two SME-ASGDs are
equivalent when the objective functions are quadratic. We remark that the presented results make use of a few simplifying approximations which are made in a non-rigorous and non-quantified manner, e.g, assuming the noise coefficients to be constant $\sigma$ and the accumulation of i.i.d noise.

Section 3 provides some numerical analysis for SME-ASGD by providing a strong approximation
estimation to the ASGD algorithm. Different from the usual convergence studies, we do not assume
convexity on $f$ or $f_i$ but only require their gradients to be (uniformly) Lipschitz. Numerical
results including non-linear forcing terms and non-convex objectives demonstrate that SME-ASGD
provides much more accurate predictions for the behavior of ASGD compared to SME-SGD derived
in~\cite{li2015stochastic}. In Section 4, we apply the optimal control theory to identify the
optimal mini-batch for ASGD and the numerical simulations there verify that the suggested strategy
gives a significantly better performance.

\section{Stochastic Modified Equations}


The asynchronous stochastic gradient descent (ASGD) carries out the following update at each step:
\begin{align}\label{ASGD}
  x_{k+1} = x_k - \eta \nabla f_{\gamma_k} (x_{k-\tau_k}),
\end{align}
where $\eta$ is the step size, $\{\gamma_k\}$ are i.i.d.~uniform
random variables taking values in $\{1, 2, \cdots , n\}$, and
$x_{k-\tau_k}$ is the delayed read of the parameter $x$ used to update
$x_{k+1}$ with a random {\em staleness} $\tau_k$. 

\smallskip 
\begin{assump}\label{assump}
  We assume that the staleness $\tau_k$ are independent and that the sample selection process
  $\gamma_k$ is mutually independent from the staleness process $\tau_k$. $\nabla f_i$'s are all
  (uniformly) Lipschitz, that is, for each $1\leq i\leq n$, there exists $L_i>0$ such that for any
  $x, y \in \mathbb{R}^d$, we have $|\nabla f_i(x) - \nabla f_i(y)|\leq L_i|x-y|$. As a consequence,
  by taking $L = \frac{1}{n}\sum_{i=1}^n L_i$, $\nabla f$ is also (uniformly) Lipschitz: $|\nabla
  f(x)-\nabla f(y)|\leq L |x-y|$. In addition, the staleness process $\tau_k$ follows the geometric
  distribution: $\tau_k = l$ (i.e., $x_{k-\tau_k} = x_{k-l}$), $l\in \{0,1,2,\cdots\}$, with probability $(1-\mu)\mu^{l}$ for
  $\mu \in (0,1)$.
\end{assump}
\begin{assump}
 We assume that the equations SME-ASGD  \eqref{linearASGD} and SME-MSGD \eqref{mosme} are ergodic.
\end{assump}

The geometric distribution assumption here is not only made to simplify the computation, but also
can be justified by considering the canonical queuing model \cite{younes2005verification}. For
example, the computation at each processor may involve a randomized algorithm that requires each
processor to do multiple independent trials until the result is accepted, thus resulting in a
geometrically distributed computation time. The geometric staleness assumption has been used in the previous asynchrony analysis, for example, see \cite{mitliagkas2016asynchrony}. Our derivation of SME models can be also easily
generalized to other random staleness models if the memory kernel, i.e., the distribution of staleness in time, decays sufficiently fast for integrability and is completely monotone when we approximate the memory kernel by a $C^{\infty}(0,\infty)$ function $\kappa(r)$. $\kappa(r)$ is completely monotone if for all for $n\geq 0, ~ r>0,~ (-1)^n\frac{d^{n}}{dr^n}\kappa(r)\geq 0.$
Under that circumstance, we can approximate the kernel accurately by $\sum_{k=1}^{n_k} c_k e^{-\lambda_k r}$ using the Bernstein's theorem of monotone functions \cite{bernstein1929}, 
and each term can be embedded into one auxiliary value to derive the SME formulation.

\subsection{Linear gradients}
We first show the derivation of Langevin dynamics with the linear forcing term. Suppose that, for
each $1\leq i\leq n, \nabla f_i$ is linear, or equivalently each $f_i$ is quadratic. While this is a
fairly restrictive assumption, the derivation in this simplified scenario offers a more transparent
view towards the stochastic modified equation for the asynchronous algorithm.

A key quantity for our derivation is the {\em expected read} $m_k$ defined as the expectation of 
$x_k$ following Assumption \ref{assump}:
\begin{equation*}
  m_k = \mathbb{E}_{\tau}( x_{k-\tau_k} )= \sum\nolimits_{l=0}^{\infty}x_{k-l} (1-\mu) \mu^{l}.
\end{equation*}
Here $m_k$ is a conditional expectation conditioned on the history of $x$, and $m_k$ is random since $x_{k-l}$'s are. Note that $m_{k+1} = \sum_{l=0}^{\infty}x_{k+1-l} (1-\mu) \mu^{l} = x_{k+1} (1-\mu) + \mu m_k$ and
$x_{k+1}= (m_{k+1}-\mu m_k)/(1-\mu)$. Plugging this into \eqref{ASGD}, we can rewrite ASGD as
\begin{align}\label{seclinear}
  \frac{m_{k+1} - 2m_k + m_{k-1}}{\eta(1-\mu)} = -\frac{m_k - m_{k-1}}{\eta} - \nabla f(m_k) + (\nabla
  f(m_k) - \nabla f_{\gamma_k}(x_{k-\tau_k})).
\end{align}
The left hand side and the first term on the
right hand side of \eqref{seclinear} can be viewed as divided difference approximations to various
time derivatives of $m$. The second term on the right hand side is the usual gradient. The last term
$\nabla f(m_k) - \nabla f_{\gamma_k}(x_{k-\tau_k})$ can be understood as the noise due to stochastic gradient
and the read delays; it has mean $0$, since the expectation, conditioned on the history of updates, can be decomposed as
\begin{multline*}
  \mathbb{E}_{\gamma,\tau}\bigl(\nabla f(m_k) -\nabla f_{\gamma_k}(m_k) + \nabla f_{\gamma_k}(m_k)- \nabla
    f_{\gamma_k}(x_{k-\tau_k}) \bigr) =\frac{1}{n}\sum_{i=1}^n (\nabla f(m_k)- \nabla f_i(m_k)) \\ +
  \frac{1}{n}\sum_{i=1}^n \big(\nabla f_i(\sum_{l=0}^{\infty}x_{k-l} (1-\mu) \mu^{l}) -
  \sum_{m=0}^{\infty}(1-\mu)\mu^m \nabla f_i(x_{k-m})\big) = 0.
\end{multline*}
The covariance matrix of the noise will be denoted as
\begin{equation*}
  \Sigma_k = \mathbb{E}_{\gamma,\tau}\bigl((\nabla f(m_k) - \nabla f_{\gamma_k}(x_{k-\tau_k}))(\nabla f(m_k) - \nabla
    f_{\gamma_k}(x_{k-\tau_k}))^T\bigr),
\end{equation*}
conditioned on $\{x_{k-l}\}_{l\ge 0}$ and we also denote the square root of $\Sigma_k$ by
$\sigma_k$, i.e., $\Sigma_k =\sigma_k\sigma_k^T$.  $\Sigma_k$ (and thus $\sigma_k$) in general
depends on the previous history of the trajectory, although such dependence is omitted in our
notation.


In order to arrive at a continuous time stochastic modified equation from \eqref{seclinear}, we view
$m_k$ as the evaluation of a function $m$ at time points $t_k = k \Delta t$ where $\Delta t$ is the 
effective time step size for the corresponding stochastic modified equation, and it is chosen as 
$\Delta t = \sqrt{\eta ( 1 - \mu)}$. By introducing the auxiliary variable 
$p_{k} = \frac{1}{\Delta t} (m_k - m_{k-1})$, we can reformulate \eqref{seclinear} as a system
of $(m_k, p_k)$:
\begin{align}\label{part1}
    p_{k+1} & = p_k -  \Delta t \sqrt{(1-\mu)/\eta} p_k - \Delta t   \nabla f(m_k) + 
    \Delta t \, \bigl(\nabla f(m_k) - \nabla f_{\gamma_k}(x_{k-\tau_k})\bigr),\\
    \label{part2}
    m_{k+1} & = m_k + \Delta t \, p_{k+1}. 
\end{align}
To obtain an SME, we first model the random term by a Gaussian random noise, that is, $\Delta t
\bigl(\nabla f(m_k) - \nabla f_{\gamma_k}(x_{k-\tau_k})\bigr) \sim \sigma_k (\eta (1-\mu))^{1/4} \Delta B_t$,
where $\Delta B_t = B_{t + \Delta t} - B_t$ is the increment of a Brownian motion (thus
$\mathbb{E}(\Delta B_t) = 0$ and $\mathbb{E} (\Delta B_t\Delta B_t^T) = \Delta t$) and the coefficient is
chosen to match the variance. Such modelling is valid because the random variables $\gamma_k$ and $\tau_k$ are independent to each other, and the choices are independent at each iteration, we can approximate the i.i.d random random term by Gaussian noise in the weak sense. Assuming that $\Delta t$ is small, we arrive at a Langevin type
equation:
\begin{equation}\label{linearASGD}
  \begin{aligned}
    dP_t & = - \nabla f (M_t) dt-\sqrt{(1-\mu)/\eta} P_t dt + \sigma(t)(\eta (1-\mu))^{1/4} dB_t,\\
    dM_t & = P_t dt,
  \end{aligned}
\end{equation}
where $\Sigma(t) = \Sigma(\{M_s\}_{0\leq s<t}, \{P_s\}_{0\leq s<t})$ has the evolution equation (the derivation is deferred to Appendix A)
\begin{align*}
    d\Sigma_t =& -\sqrt{\frac{1-\mu}{\eta}} \Sigma_t dt -\mu (\nabla f(M_t)\nabla f(P_t)^T +\nabla f(P_t)\nabla f(M_t)^T) dt - \sqrt{\frac{1-\mu}{\eta}}\nabla f(M_t)\nabla f(M_t)^T dt\\
    &+\mu \sqrt{\eta(1-\mu)} \nabla f(P_t)\nabla f(P_t)^T dt + \frac{1}{n}\sqrt{\frac{1-\mu}{\eta}}\sum_{i=1}^{n}\nabla f_i(M_t + \mu\sqrt{\frac{\eta}{1-\mu}} P_t)\nabla f_i(M_t + \mu\sqrt{\frac{\eta}{1-\mu}} P_t)^T dt.
\end{align*}

When $f$ is a smooth confining potential, that is, $f$ satisfies $\lim_{|x|\to +\infty} f(x) = +\infty$ and $e^{-\beta f(x)}\in L^1(\mathbb{R}^d)$ for all $\beta\in \mathbb{R}^+$ (an example for $f$ is being a quadratic potential), the process
approaches to the minimum of the potential function, and $\sigma(t)$ (as the damping term $-\sqrt{(1-\mu)/\eta} \Sigma_t dt$ dominates in the evolution equation) can be approximated by a constant matrix $\sigma$ up to a first order approximation for large time $t$. When this constant matrix $\sigma$ is a multiple of the identity
matrix, say $\sigma = \varsigma I$, $(P_t, M_t)$ in the standardized model is an ergodic Markov process with stationary
distribution \cite{pavliotis2014stochastic}:
\[
  \rho_{\infty}(p,m) = Z^{-1}  e^{-\beta (\frac{1}{2} |p|^2 + f(m)) },
\]
where $Z$ is a normalization constant. In this case, the resulting friction is
$\sqrt{(1-\mu)/\eta}$ and the temperature $\beta^{-1}$ is $\frac{1}{2} \varsigma^2 \eta$.
When the constant matrix $\sigma$ is not a multiple of identity (but still being constant), the stationary 
distribution takes a similar form in a
transformed coordinate system. We remark that though in theory proving time-inhomogeneous process (\ref{linearASGD}) has a unique stationary distribution is beyond the scope of this paper, the numerical observations suggest that such a constant approximation of the noise coefficient does not change the process' property fundamentally; in the numerical experiments, we observe that the trajectory of SME-ASGD does not change much when we replace the coefficient  of  noise by a constant matrix. 

The reason why we care about the temperature parameter here is that it quantifies the variance of
the noise and therefore gives us more information about the asymptotic behavior of the optimization
process. With such a tool, we can better analyze the connection between different stochastic
gradient algorithms. Let us illustrate it by showing one example here: Mitliagkas et
al. \cite{mitliagkas2016asynchrony} argues that there is some equivalence between adding asynchrony
or momentum to the SGD algorithms, and they showed it by taking expectation to a simple queuing
model and finding matched coefficients. Here, we investigate such relation by looking at the
corresponding Langevin dynamics, specifically the temperature for both SMEs, thus offering a more
detailed dynamical comparison.

Stochastic gradient descent with momentum (MSGD) introduced by \cite{polyak1964some} utilizes the
velocity vector from the past updates to accelerate the gradient descent
\cite{sutskever2013importance}:
\begin{equation}\label{momentum}
  \begin{aligned}
    & v_{k+1} = \mu' \, v_k - \eta' \nabla f_{\gamma_k}(x_k), \\
    & x_{k+1} = x_k + v_{k+1},
  \end{aligned}
\end{equation}
with a momentum parameter $\mu' \in (0,1)$. \eqref{momentum} can be also viewed as a discretization
of a second-order stochastic differential equation. Our derivation here is slightly different from 
\cite{li2015stochastic} since we use a more natural time scale $\Delta t =\sqrt{\eta'}$ in order to 
obtain an SDE with bounded coefficients. By taking $p$  to be $v/\sqrt{\eta'}$ (see Appendix A), we end
up with the following stochastic modified equation for MSGD (denoted in short as SME-MSGD)
\begin{align}\label{mosme}
& dP_t = -\nabla  f(X_t) dt - \frac{1-\mu'}{\sqrt{\eta'}}P_t dt + \sigma(X_t)(\eta')^{\frac{1}{4}} dB_t, \nonumber \\
& dX_t = P_t dt,
\end{align}
where the friction is $\frac{1-\mu' }{\sqrt{\eta'}}$. 
Note that \eqref{mosme} is time-homogeneous with an multiplicative noise, such that the invariant measure usually does not have an explicit expression in general. We further postulate that when the noise is small, the coefficient $\sigma(X_t)$ can be approximated by a constant multiple of the identity matrix. In this case, the temperature ${\beta'}^{-1} =
\frac{\varsigma^2 \eta'}{2(1-\mu' )}$ dictates the convergence rate to the stationary solution. If we further assume that the noise coefficients $\sigma$ in SME-ASGD  \eqref{linearASGD} and in SME-MSGD \eqref{mosme} are the same constant, comparing
\eqref{linearASGD} with \eqref{mosme} results in the following interesting observation.

\smallskip
\begin{proposition}
If we assume that the noise coefficients $\sigma$ in SME-ASGD  \eqref{linearASGD} and in SME-MSGD \eqref{mosme} are the same constant, if $\mu' = \mu$ and $\eta' = \eta (1-\mu)$, then \eqref{linearASGD} and \eqref{mosme} have the same stationary distribution. 
\end{proposition}

In Theorems 3 and 5 in Mitliagkas et al.'s paper \cite{mitliagkas2016asynchrony}, the staleness'
geometric distribution parameter $\mu $ is taken to be $ \mu' = 1-\frac{1}{M}$, where $M$ is
the number of mutually independent workers and $\mu'$ is the momentum parameter. With these
assumptions,  when looking at \eqref{linearASGD} and \eqref{mosme} under the same time scale with 
$\eta' = \eta (1-\mu)$, we can see that  ${\beta'}^{-1}  = \frac{\varsigma^2 \eta'}{2(1-\mu' )} 
= \frac{\varsigma^2 \eta}{2} = \beta^{-1}$. Since the corresponding temperature for the asynchronous method and momentum method are equal, we conclude that the
perspective of stochastic modified equation given above explains the observation in
\cite{mitliagkas2016asynchrony} that the momentum method has certain equivalent performance as the
asynchronous method.

\subsection{Nonlinear gradients}
We now consider the general case in which the gradient $\nabla f_i$
can be non-linear. One can still write the ASGD into a stochastic
modified equation.
For this, let us define a new auxiliary variable $y_k$ which is proportional to 
the expected gradient:
\begin{align}\label{newsum}
  y_k = -\alpha \mathbb{E}_{\tau} (\nabla f(x_{k-\tau_k}) )= -\alpha \sum\nolimits_{l=0}^{\infty} \nabla f(x_{k-l})
  (1-\mu)\mu^l,
\end{align}
where $\alpha>0$ is to be determined. Again $y_k$ is random and a conditional expectation conditioned on the history of $x$.
Directly following the definition, $y_k$ satisfies the difference equation
\begin{align}\label{ypart}
  \frac{y_{k+1}-y_k}{\alpha(1-\mu)} = -\frac{y_k}{\alpha}-\nabla f(x_{k+1}).
\end{align}
Moreover, we can rewrite the ASGD \eqref{ASGD} as
\begin{align}\label{xpart}
  \frac{x_{k+1}-x_k}{\eta/\alpha} = y_k + \alpha \Bigl(-\frac{y_k}{\alpha} - \nabla f_{\gamma_k}
  (x_{k-\tau_k}) \Bigr).
\end{align}
The reason for us arranging terms in this way is to formulate a Langevin-type equation, but with the
noise term moved from the momentum side ($Y$) to the position side ($X$). Notice that on the right
hand side of \eqref{xpart}, $-\frac{y_k}{\alpha} - \nabla f_{\gamma_k} (x_{k-\tau_k})$ can be viewed
as a noise with mean $0$
\begin{align*}
  \mathbb{E}_{\gamma,\tau}\Bigl(-\frac{y_k}{\alpha} - \nabla f_{\gamma_k} (x_{k-\tau_k})\Bigr) &= \frac{1}{n}\sum_{i=1}^n \mathbb{E}_{\tau}\bigg( \sum_{l=0}^{\infty} \nabla f(x_{k-l}) (1-\mu)\mu^l - \nabla f_i (x_{k-\tau_k})\bigg)\\
  &=\mathbb{E}_{\tau}\bigg( \sum_{l=0}^{\infty} \nabla f(x_{k-l}) (1-\mu)\mu^l - \nabla f (x_{k-\tau_k})\bigg)\\
  &=\sum_{m=0}^{\infty} (1-\mu)\mu^m \big(\sum_{l=0}^{\infty} \nabla f(x_{k-l}) (1-\mu)\mu^l - \nabla f (x_{k-m})\big) \\
  &=\sum_{l=0}^{\infty} \nabla f(x_{k-l}) (1-\mu)\mu^l-\sum_{m=0}^{\infty}\nabla f (x_{k-m}) (1-\mu)\mu^m = 0.
\end{align*}
And the covariance matrix conditioned on $x_{k-l}, l=0,1,2,\cdots$ is given by
\begin{align*}
  \Sigma_k &= \frac{1}{n}\sum_{i=1}^n
  \mathbb{E}\left(\left(-\frac{y_k}{\alpha} - \nabla f_i(x_{k-\tau_k})\right)
  \left(-\frac{y_k}{\alpha} - \nabla f_i(x_{k-\tau_k})\right)^T\right)\\
  & = \frac{1}{n}\sum_{i=1}^n \mathbb{E}\left(\left(\sum_{l=0}^{\infty} \nabla f(x_{k-l}) (1-\mu)\mu^l  -
  \nabla f_i(x_{k-\tau_k})\right)
  \left(\sum_{l=0}^{\infty} \nabla f(x_{k-l}) (1-\mu)\mu^l - \nabla f_i(x_{k-\tau_k})\right)^T\right).
\end{align*}
In order to view \eqref{ypart} and \eqref{xpart} as a time-discretization of a coupled system with
the same time step size, we match $\alpha(1-\mu)$ with $\eta/\alpha$ by choosing
$\alpha = \sqrt{\nicefrac{\eta}{(1-\mu)}}$.  
Setting the step size $\Delta t=\alpha(1-\mu)=\eta/\alpha=\sqrt{\eta(1-\mu)}$
and taking a Gaussian approximation to the noise
$\eta\big(-\frac{y_k}{\alpha}-\nabla f_{\gamma_k} (x_{k-\tau_k}) \big)\sim
\sqrt{\Sigma_k}\frac{\eta^{3/4}}{(1-\mu)^{1/4}} \Delta B_t$, we arrive at the stochastic modified
equation for the nonlinear case
\begin{equation}\label{nonlinearASGD}
\begin{aligned}
dY_t &= -\nabla f(X_t) dt -\sqrt{\frac{1-\mu}{\eta}} Y_t dt \\
dX_t &= Y_t dt + \sqrt{\Sigma(t)}\frac{\eta^{3/4}}{(1-\mu)^{1/4}} dB_t
\end{aligned}
\end{equation}
Here $\Sigma(t) = \Sigma(\{X_s\}_{0\le s<t},\{Y_s\}_{0\le s<t})$. In order to close the system of
equations, we derive an explicit evolution equation for $\Sigma$
\begin{multline}\label{evol}
  d\Sigma_t = -\sqrt{\frac{1-\mu}{\eta}}\Sigma_t dt + \sqrt{\frac{1-\mu}{\eta}}\Bigl(
  \frac{1}{n}\sum_{i=1}^{n} \nabla f_i(X_t) \nabla f_i(X_t)^T +\frac{1-\mu}{\mu} \nabla f(X_t)
  \nabla f(X_t)^T \Bigr) dt \\ +\frac{1-\mu}{\eta \mu} \bigl(\sqrt{\frac{1-\mu}{\eta}}Y_t Y_t^T
  +\nabla f(X_t) Y_t^T + Y_t \nabla f(X_t)^T\Bigr) dt.
\end{multline}
The derivation of \eqref{evol} is shown in Appendix A. The combined system
\eqref{nonlinearASGD}--\eqref{evol} will be referred as SME-ASGD (the stochastic modified equations
for asynchronous SGD) for the general nonlinear-gradient case. We should point it out that unlike
the linear-gradient case \eqref{linearASGD} , \eqref{nonlinearASGD} has no known explicit formula
for invariant measure even when $\Sigma(t)$ converging to a constant matrix. Nevertheless, the
ergodicity of \eqref{nonlinearASGD} and \eqref{evol} will be an interesting future direction to
explore.

We would like to point out that when the gradient $\nabla f$ is linear \eqref{ypart}
  and \eqref{xpart} can be easily transformed back to \eqref{part1}
  and \eqref{part2}. As a consequence, \eqref{linearASGD} and
  \eqref{nonlinearASGD} are equivalent. To see this, 
\[
y_k = -\alpha \nabla f(\sum_{l=0}^{\infty} x_{k-l} (1-\mu)\mu^l) = -\alpha \nabla f(m_k).
\]
Replacing $y_{k+1}$ and $y_k$ with the above formula and also $x_{k+1}$ with $\frac{m_{k+1}-\mu
  m_k}{1-\mu}$, we can rewrite \eqref{ypart} as
\[
-\frac{\nabla f(m_{k+1}) - \nabla f(m_k)}{1-\mu} = \nabla f(m_k) - \nabla f( \frac{m_{k+1}-\mu
  m_k}{1-\mu}) = -\frac{1}{1-\mu}\nabla f(m_{k+1} - m_k).
\]
Since $p_{k+1} = (m_{k+1} - m_k)/\sqrt{\eta(1-\mu)}$, we have
\[
\nabla f(m_{k+1}-m_k) = \nabla f(p_{k+1 }\sqrt{\eta(1-\mu)}),
\]
which implies \eqref{part2}. To show \eqref{part1}, we first notice that
\begin{align*}
  \frac{x_{k+1} - x_k}{\eta/\alpha} &= \frac{m_{k+1} - (\mu+1)m_k + \mu m_{k-1}}{(1-\mu )\eta/ \alpha}= \frac{m_{k+1} - 2m_k +m_{k-1}}{(1-\mu)\eta/\alpha} +\frac{m_k - m_{k-1}}{\eta/\alpha}\\
  & = \frac{p_{k+1} - p_k}{1-\mu} + p_k = -\alpha \nabla f(m_k) + \alpha (\nabla f(m_k) - \nabla f_{\gamma_k}(v_k))\\
  & = -\sqrt{\frac{\eta}{1-\mu}} \nabla f(m_k) + \sqrt{\frac{\eta}{1-\mu}} (\nabla f(m_k) - \nabla f_{\gamma_k}(v_k))
\end{align*}
by plugging in $\alpha$ in terms of $\mu, \eta$. It is clear now that this gives \eqref{part1}.


\section{Approximation error of the stochastic modified equation}

The difference between the time-discrete ASGD and the time-continuous SME-ASGD can be rigorously
quantified as follows.


\begin{theorem}\label{thm}
  Assume that Assumption~\ref{assump} holds and that the variance from the asynchronous
  gradients is uniformly bounded (i.e., there exists $c>0$ such that $||\sigma(t)||\leq c$). Suppose also that 
  all the iterates updated from the ASGD stay bounded and that the solutions for SME-ASGD and ASGD before time $0$
  agree (i.e., $X_{l\Delta t} = x_l, l\leq 0$, with $\Delta t = \sqrt{\eta(1-\mu)}$
  as given previously). Then the SME-ASGD approximates the ASGD in the sense that there exists constant $K_T>0$
  depending only on $T$ such that
  \begin{align}
    \sup_{n\Delta t \leq T} \mathbb{E}\big\{|X_{n \Delta t} - x_n|\big\} \leq K_T \frac{\Delta t}{1-\mu}
  \end{align}
  for $\Delta t$ sufficiently small. Here $X_{n\Delta t}\equiv X(n\Delta t)$ is the
  solution of \eqref{nonlinearASGD} at time $n\Delta t$ and $x_n$ is
  from ASGD \eqref{ASGD}.
\end{theorem}
The assumption $\sigma = \sqrt{\Sigma} = O(1)$ can be justified from \eqref{evol} as $\Sigma$ is
approximated by a constant matrix for $t$ large. This is because when the iterate approaches to the
minimizer, the gradients are close to $0$, and $Y_t$ converges to be a constant vector. Since we investigate the error approximation in finite time $T$ and finite step size $\Delta t$, there are only a finite number of iterations. In each iteration, the iterate updated from the ASGD stays bounded by a sufficient large constant with high probability. Therefore, the assumption that all iterates stay bounded by a sufficient large constant holds with high probability.

The proof of the Theorem \eqref{thm} follows from viewing the ASGD as a discretization of SME-ASGD
and using the analysis of strong convergence for numerical schemes for stochastic differential
equations (SDEs). \\
\begin{proof}[Proof of the Theorem \eqref{thm}]
We look at the one step approximation in the first step, and the global approximation can be done by
induction. Using the variation of constant formula, we know that the solution of 
\[
dY_t = -\nabla f(X_t) dt -\sqrt{\frac{1-\mu}{\eta}} Y_t dt
\]
is given by 
\[
Y_t = e^{-\sqrt{\frac{1-\mu}{\eta}}t} Y_0 -  \int_0^t e^{-\sqrt{\frac{1-\mu}{\eta}}(t-s)} \nabla f(X_s) ds,
\]
where $Y_0 = -\sqrt{\frac{\eta}{1-\mu}}\sum_{l=0}^{\infty} \nabla f(x_{-l})(1-\mu) \mu^l$ as defined
in \eqref{newsum}. Plugging $Y_t$ into the integral form of $X_{\Delta t}$ gives rise to
\begin{align}
  X_{\Delta t} = x_0 + \int_0^{\Delta t} \bigg( e^{-\sqrt{\frac{1-\mu}{\eta}}s} Y_0 -
   \int_0^s e^{-\sqrt{\frac{1-\mu}{\eta}}(s-u)}\nabla f(X_u) du \bigg)ds +
  \frac{\eta^{3/4}}{(1-\mu)^{1/4}} \int_0^{\Delta t} \sigma(s) dB_s.
\end{align}
Denote $v_k := x_{k-\tau_k}$ for notation convenience. By splitting $\eta \nabla f_{\gamma_0}(v_0)$ into $\eta \nabla f_{\gamma_0}(v_0) - \eta
\sum_{l=0}^{\infty} \nabla f(x_{-l})(1-\mu) \mu^l$ and $ \eta \sum_{l=0}^{\infty} \nabla
f(x_{-l})(1-\mu) \mu^l$, we can make the following estimate
\begin{align*}
  \mathbb{E}\big\{|X_{\Delta t} &- x_1 |\big\}\leq  \bigg| \int_0^{\Delta t}  e^{-\sqrt{\frac{1-\mu}{\eta}}s} Y_0ds + \eta \sum_{l=0}^{\infty} \nabla f(x_{-l})(1-\mu) \mu^l \bigg|  \\
  &+\mathbb{E}\bigg\{ \int_0^{\Delta t} \bigg( \int_0^s e^{-\sqrt{\frac{1-\mu}{\eta}}(s-u)}\big|\nabla f(X_u)-\nabla f(x_1) \big| du \bigg) ds \bigg\} + |\nabla f(x_1)| \int_0^{\Delta t} \int_0^s e^{-\sqrt{\frac{1-\mu}{\eta}}(s-u)}du ds\\
  &+ \frac{\eta^{3/4}}{(1-\mu)^{1/4}}\bigg( \mathbb{E}\big\{ \big(\int_0^{\Delta t} \sigma(s) dB_s \big)^2\big\}\bigg)^{1/2}+ \mathbb{E}\big\{\big|\eta \nabla f_{\gamma_0}(v_0) - \eta \sum_{l=0}^{\infty} \nabla f(x_{-l})(1-\mu) \mu^l \big| \big\} \\
  &\leq I + II + III + \frac{\eta^{3/4}}{(1-\mu)^{1/4}}\bigg( \mathbb{E}\big\{ \int_0^{\Delta t} \sigma(s)^2 ds \big\}\bigg)^{1/2} + c \eta  \leq I + II + III + 2c \frac{\Delta t^2}{1-\mu},
\end{align*}
where $I$, $II$, and $III$ are the first three terms appeared in the right hand side of the first inequality. In the above derivation, we have applied the Ito isometry to the fourth term and used
\[
\frac{\eta^{3/4}}{(1-\mu)^{1/4}}\bigg( \mathbb{E}\big\{ \int_0^{\Delta t} \sigma(s)^2 ds
\big\}\bigg)^{1/2} \leq c \frac{\Delta t^2}{1-\mu},
\]
since $\Delta t = \sqrt{\eta(1-\mu)}$. The fifth term, after an application of the Cauchy-Schwarz inequality,
is shown to be a discrete version of the covariance matrix
\begin{align*}
  \mathbb{E}\big\{\big|\eta \nabla f_{\gamma_0}(v_0) - \eta \sum_{l=0}^{\infty} \nabla f(x_{-l})(1-\mu) \mu^l \big| \big\}\leq \eta \sqrt{\Sigma_0} \leq c\eta.
\end{align*}
Let us now treat the first three terms
\begin{align*}
I &= \bigg| \int_0^{\Delta t} e^{-\sqrt{\frac{1-\mu}{\eta}}s} Y_0 ds + \eta \sum_{l=0}^{\infty} \nabla f(x_{-l})(1-\mu) \mu^l \bigg|\\
&=\bigg| \sqrt{\frac{\eta}{1-\mu}} (e^{-\sqrt{\frac{1-\mu}{\eta}}\Delta t} -1) \sqrt{\frac{\eta}{1-\mu}}\sum_{l=0}^{\infty} \nabla f(x_{-l})(1-\mu) \mu^l +\eta \sum_{l=0}^{\infty} \nabla f(x_{-l})(1-\mu) \mu^l \bigg|\\
&= \bigg|- \sqrt{\frac{\eta}{1-\mu}}\sum_{l=0}^{\infty} \nabla f(x_{-l})(1-\mu) \mu^l \Delta t  + \eta \sum_{l=0}^{\infty} \nabla f(x_{-l})(1-\mu) \mu^l +O(\Delta t^2) \bigg|= O(\Delta t^2),
\end{align*}
since the first two terms cancel. Because $\nabla f$ is Lipschitz and $e^{-\sqrt{\frac{1-\mu}{\eta}}(s-u)} \le 1$ for $u\leq s$, the second term
can be estimated with
\begin{align*}
  II \leq L\Delta t \int_0^{\Delta t} \mathbb{E}\left\{|X_u - x_1| \right\} du.
\end{align*} 
Since $x_1$ stays in a bounded domain, the third term can be bounded by 
\begin{align*}
  III \leq   |\nabla f(x_1)| \int_0^{\Delta t}  s ds=  |\nabla f(x_1)| \Delta t^2 /2  = O(\Delta t^2).
\end{align*}
With these estimates available, we can choose a sufficiently large
constant $C$ (depending on $c$ and the size of the domain containing
the iterates from ASGD) such that
\begin{align*}
  \mathbb{E}\big\{|X_{\Delta t} - x_1 |\big\}&\leq C\frac{\Delta t^2}{1-\mu} + L\Delta t \int_0^{\Delta t}
  \mathbb{E}\big\{|X_u - x_1|\big\} du.
\end{align*}
An application of Gronwall's inequality shows that
\begin{align*}
  \mathbb{E}\big\{|X_{\Delta t} - x_1 |\big\}&\leq C\frac{\Delta t^2}{1-\mu}e^{L\Delta t^2}= C \frac{\Delta t^2}{1-\mu}+O(\Delta t^4)\leq C\frac{\Delta t^2}{1-\mu}.
\end{align*}
This concludes the estimate for the first step at time 0.

The induction step is similar. We have
\begin{align*}
X_{(k+1)\Delta t} = X_{k\Delta t} + \int_{k\Delta t}^{(k+1)\Delta t} \bigg(e^{-\sqrt{\frac{1-\mu}{\eta}}(s-k\Delta t)}Y_{k\Delta t}& -\int_{k\Delta t}^s e^{-\sqrt{\frac{1-\mu}{\eta}}(s-u)}\nabla f(X_u) du \bigg) ds   \\
+& \frac{\eta^{3/4}}{(1-\mu)^{1/4}} \int_{k\Delta t}^{(k+1)\Delta t} \sigma(s) dB_s.
\end{align*}
For the discrete update step $x_{k+1} = x_k -\eta \nabla f_{\gamma_k}(v_k)$, we split $\eta \nabla f_{\gamma_k}(v_k)$ 
as before. With the assumption $\mathbb{E}\big\{|X_{k\Delta t} - x_k |\big\}\leq
Ck\frac{\Delta t^2}{1-\mu}$, we have the following estimate
\begin{align*}
  \mathbb{E}\big\{|X_{(k+1)\Delta t} &- x_{k+1} |\big\}\leq \mathbb{E}\big\{|X_{k\Delta t} - x_k |\big\}+ \int_{k\Delta t}^{(k+1)\Delta t} e^{-\sqrt{\frac{1-\mu}{\eta}}(s-k\Delta t)}\big|Y_{k\Delta t} - y_k \big| ds\\
  &+\bigg| \int_{k\Delta t}^{(k+1)\Delta t}  e^{-\sqrt{\frac{1-\mu}{\eta}}(s-k\Delta t)}  y_k ds + \eta \sum_{l=0}^{\infty} \nabla f(x_{k-l})(1-\mu) \mu^l \bigg|  \\
  &+\mathbb{E}\bigg\{ \int_{k\Delta t}^{(k+1)\Delta t} \bigg( \int_{k\Delta t}^s e^{-\sqrt{\frac{1-\mu}{\eta}}(s-u)}\big|\nabla f(X_u)-\nabla f(x_{k+1}) \big| du \bigg) ds \bigg\}\\
  & + |\nabla f(x_{k+1})| \int_{k\Delta t}^{(k+1)\Delta t}\int_{k\Delta t}^s e^{-\sqrt{\frac{1-\mu}{\eta}}(s-u)}du ds + \frac{\eta^{3/4}}{(1-\mu)^{1/4}}\bigg( \mathbb{E}\big\{ \big(\int_{k\Delta t}^{(k+1)\Delta t} \sigma(s) dB_s \big)^2\big\}\bigg)^{1/2}\\
  &+ \mathbb{E}\big\{\big|\eta \nabla f_{\gamma_k}(v_k) - \eta \sum_{l=0}^{\infty} \nabla f(x_{k-l})(1-\mu) \mu^l \big| \big\}.
\end{align*}
Here the only difference compared to the first step is the term $Y_{k\Delta t}$, which is not given but 
generated from SME. Note that 
\[
y_k = -\sqrt{\frac{\eta}{1-\mu}}\sum_{l=0}^{\infty} \nabla f(x_{k-l})(1-\mu) \mu^l.
\]
From \eqref{ypart}, we observe that $y_k$ is indeed an approximation of $Y_t$ by applying
the Euler discretization to the ordinary differential equation part of the SME. Because the global
truncation error for the Euler method in ODE is $O(\Delta t)$, we have
\begin{align*}
\int_{k\Delta t}^{(k+1)\Delta t}  e^{-\sqrt{\frac{1-\mu}{\eta}}(s-k\Delta t)} \big|Y_{k\Delta t} -y_k\big|ds= O(\Delta t ^2). 
\end{align*}
The third term has the estimate
\begin{align*}
 \bigg| \int_{k\Delta t}^{(k+1)\Delta t} & e^{-\sqrt{\frac{1-\mu}{\eta}}(s-k\Delta t)}  y_k ds + \eta \sum_{l=0}^{\infty} \nabla f(x_{k-l})(1-\mu) \mu^l \bigg| \\
 &= \bigg| -\sqrt{\frac{\eta}{1-\mu}}\big(e^{-\sqrt{\frac{\eta}{1-\mu}}\Delta t}-1\big)y_k + \eta \sum_{l=0}^{\infty} \nabla f(x_{k-l})(1-\mu) \mu^l \bigg|\\
 & = \bigg|-\sqrt{\frac{\eta}{1-\mu}}\sum_{l=0}^{\infty} \nabla f(x_{k-l})(1-\mu) \mu^l \Delta t + \eta \sum_{l=0}^{\infty} \nabla f(x_{k-l})(1-\mu) \mu^l  + O(\Delta t^2) \bigg|=O(\Delta t^2)
\end{align*}
as before. All other terms have the same estimates as in the base case. Applying the Gronwall's inequality again and letting $\Delta t $ be sufficiently small gives the estimate
\begin{align*}
  \mathbb{E}&\big\{\big|X_{(k+1)\Delta t} - x_{k+1} \big|\big\}\leq Ck\frac{\Delta t^2}{1-\mu}.
\end{align*}
As $n\Delta t \leq T$ for all $n$, one can conclude that there exists $K_T>0$ such that
\begin{align*}
  \mathbb{E}\big\{\big|X_{n \Delta t} - x_n \big|\big\}\leq K_T \frac{\Delta t}{1-\mu}.
\end{align*}
\end{proof}

One interesting observation is that, contrary to the standard Euler-Maruyama
method for SDEs having strong order of convergence $1/2$ \cite{kloeden1992stochastic}, the above
result indicates that ASGD, viewed as a discretization of SME-ASGD, has strong order $1$. This is
because the coefficient of the noise term in the SME-ASGD has
$\nicefrac{\eta^{3/4}}{(1-\mu)^{1/4}}$, which is of order $o(1)$. The SME model proposed in
\cite{li2015stochastic} has the same feature: the coefficient of the noise term there is of order
$\sqrt{\eta}$. When $\eta \approx 1-\mu$, the two orders are the same.

\begin{figure}[h!]
    \centering
    \begin{minipage}{0.32\textwidth}
        \centering
        \includegraphics[width=0.95\textwidth]{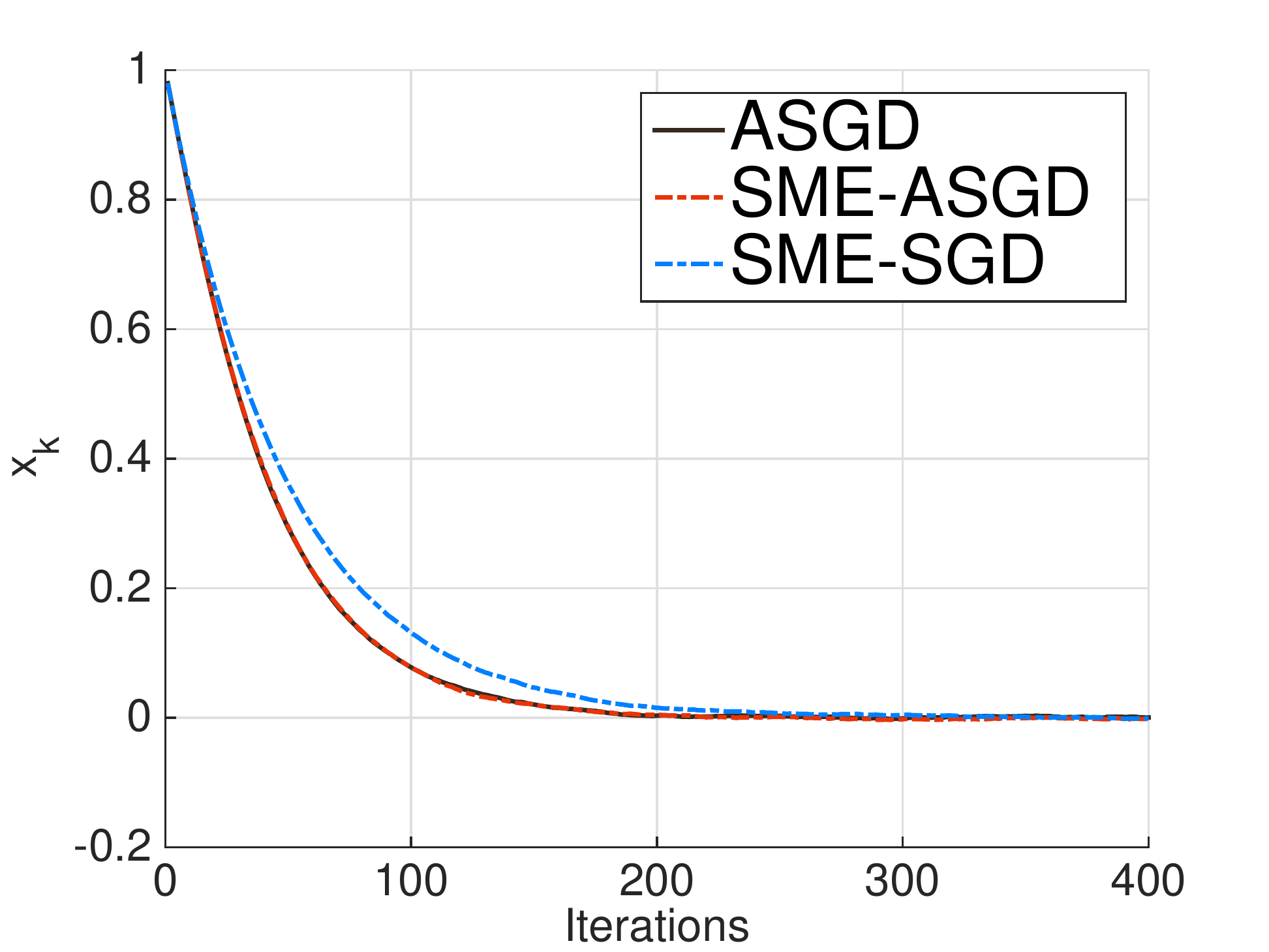} 
    \end{minipage}\hfill
    \begin{minipage}{0.32\textwidth}
        \centering
        \includegraphics[width=0.95\textwidth]{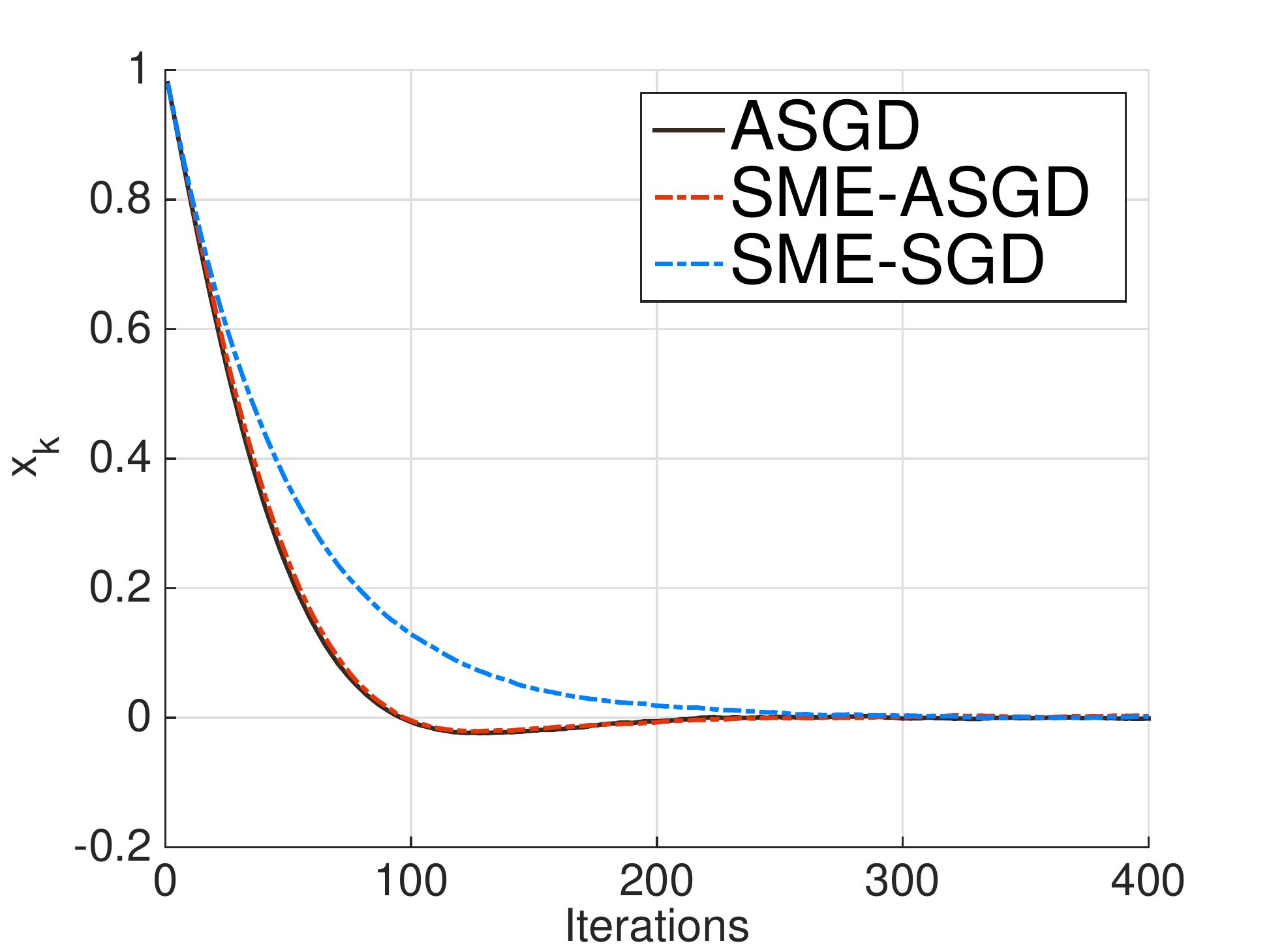} 
    \end{minipage}
        \begin{minipage}{0.32\textwidth}
        \centering
        \includegraphics[width=0.95\textwidth]{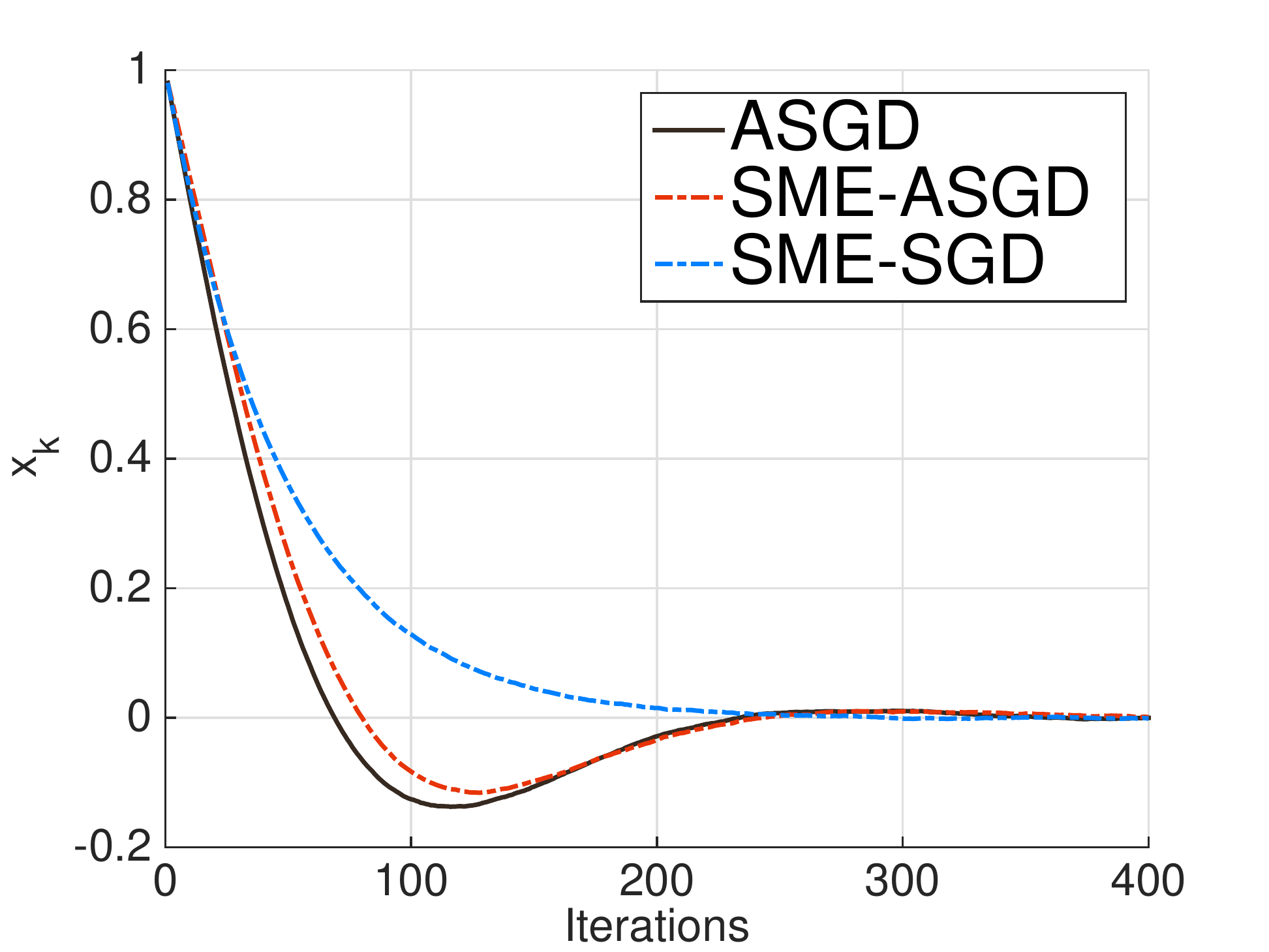} 
    \end{minipage}
    \begin{minipage}[a]{0.32\textwidth}
        \centering
        \includegraphics[width=0.95\textwidth]{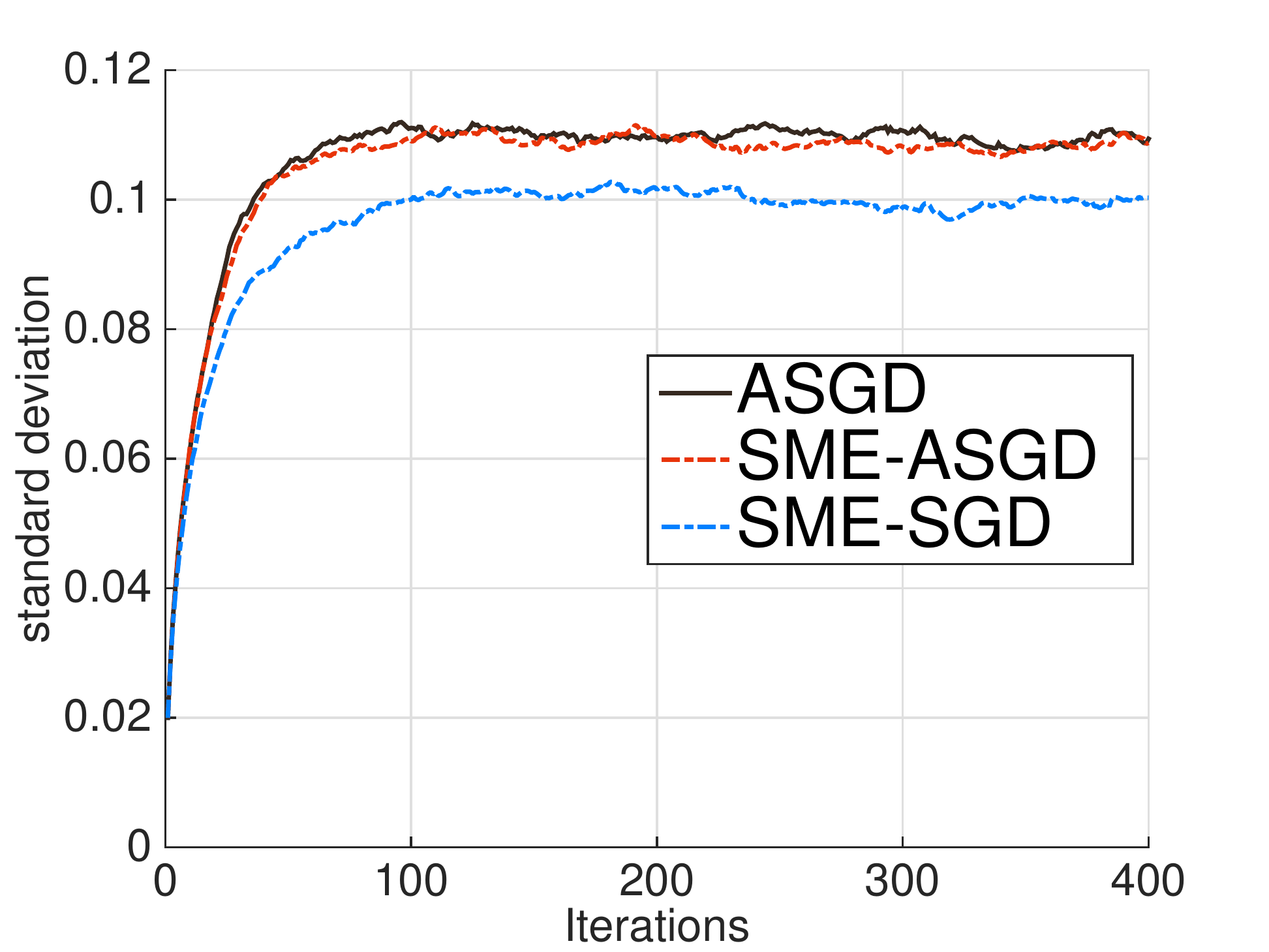} 
       \caption*%
            {{\small $\mu=0.9$}}  
    \end{minipage}\hfill
        \begin{minipage}{0.32\textwidth}
        \centering
        \includegraphics[width=0.95\textwidth]{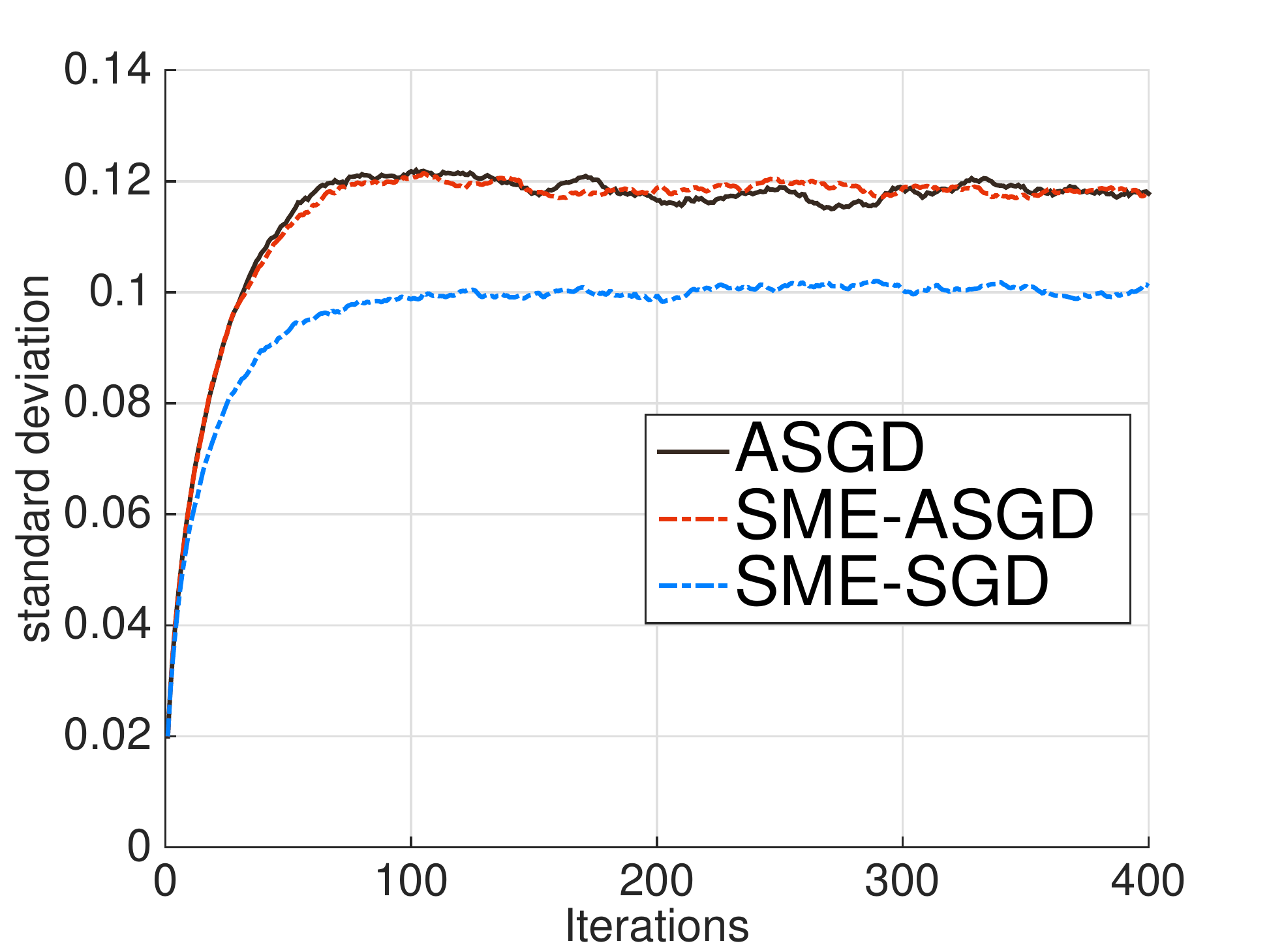} 
       \caption*%
            {{\small $\mu=0.95$}}  
    \end{minipage}
    \begin{minipage}{0.32\textwidth}
        \centering
        \includegraphics[width=0.95\textwidth]{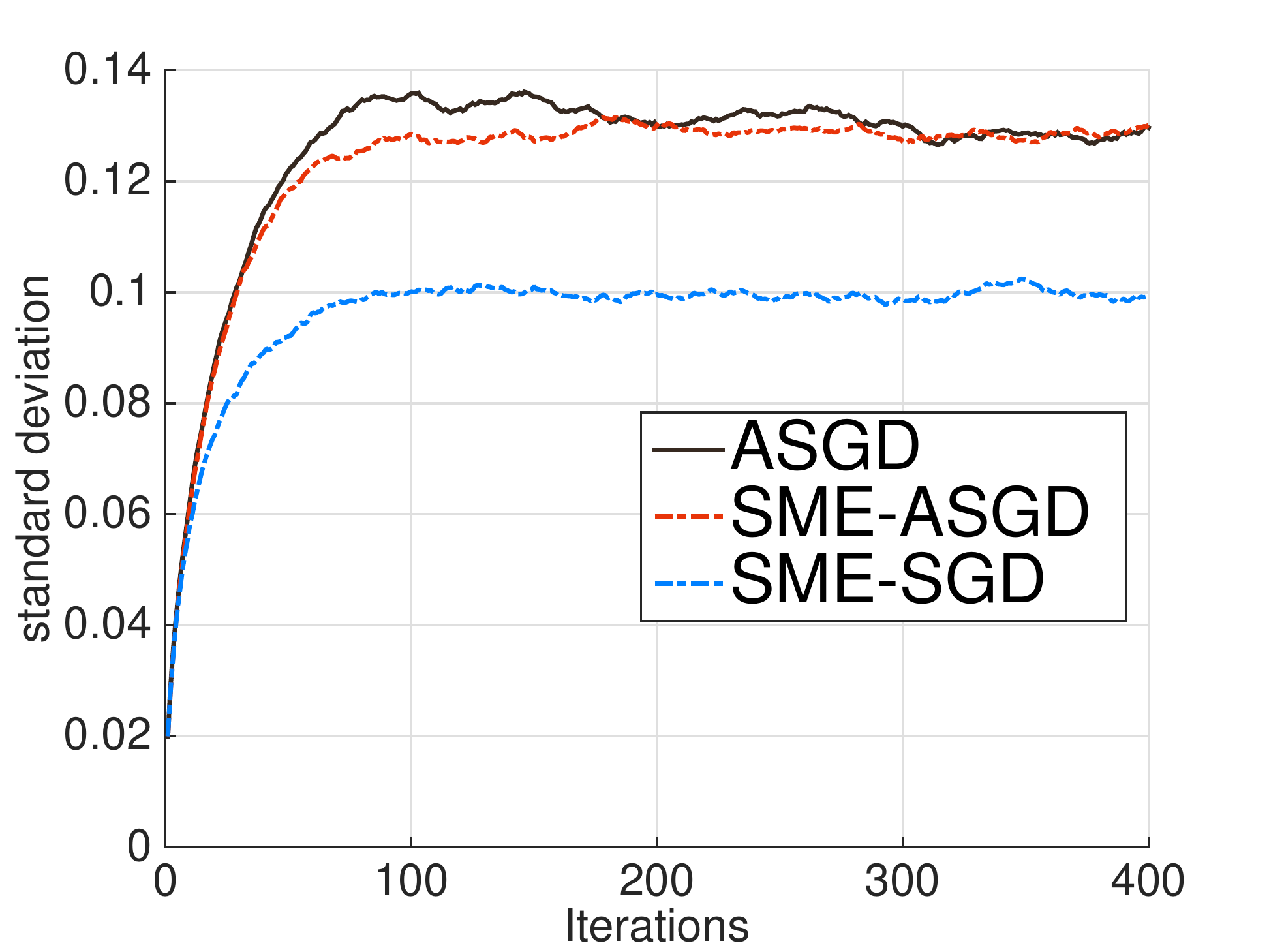} 
               \caption*%
            {{\small $\mu=0.97$}}  
    \end{minipage}
     \caption{ Apply the SME-ASGD to minimize the quadratic function $f(x) = x^2$ in different
       $\mu$'s, with two components $f_1(x) = (x-1)^2-1$, and $f_2(x) = (x+1)^2-1$.  $x_0 =1$ and
       $\eta = 1e-2$. SME-ASGD achieves more accurate approximations compared to SME-SGD
       \eqref{LiSME}, especially when $\mu$ becomes large. However, one can also observe that when
       $\mu$ increases the error of the SME-ASGD approximation increases as well.}
     \label{fig:linear}
\end{figure}

\begin{figure}[h!]
    \centering
    \begin{minipage}{0.4\textwidth}
        \centering
        \includegraphics[width=0.95\textwidth]{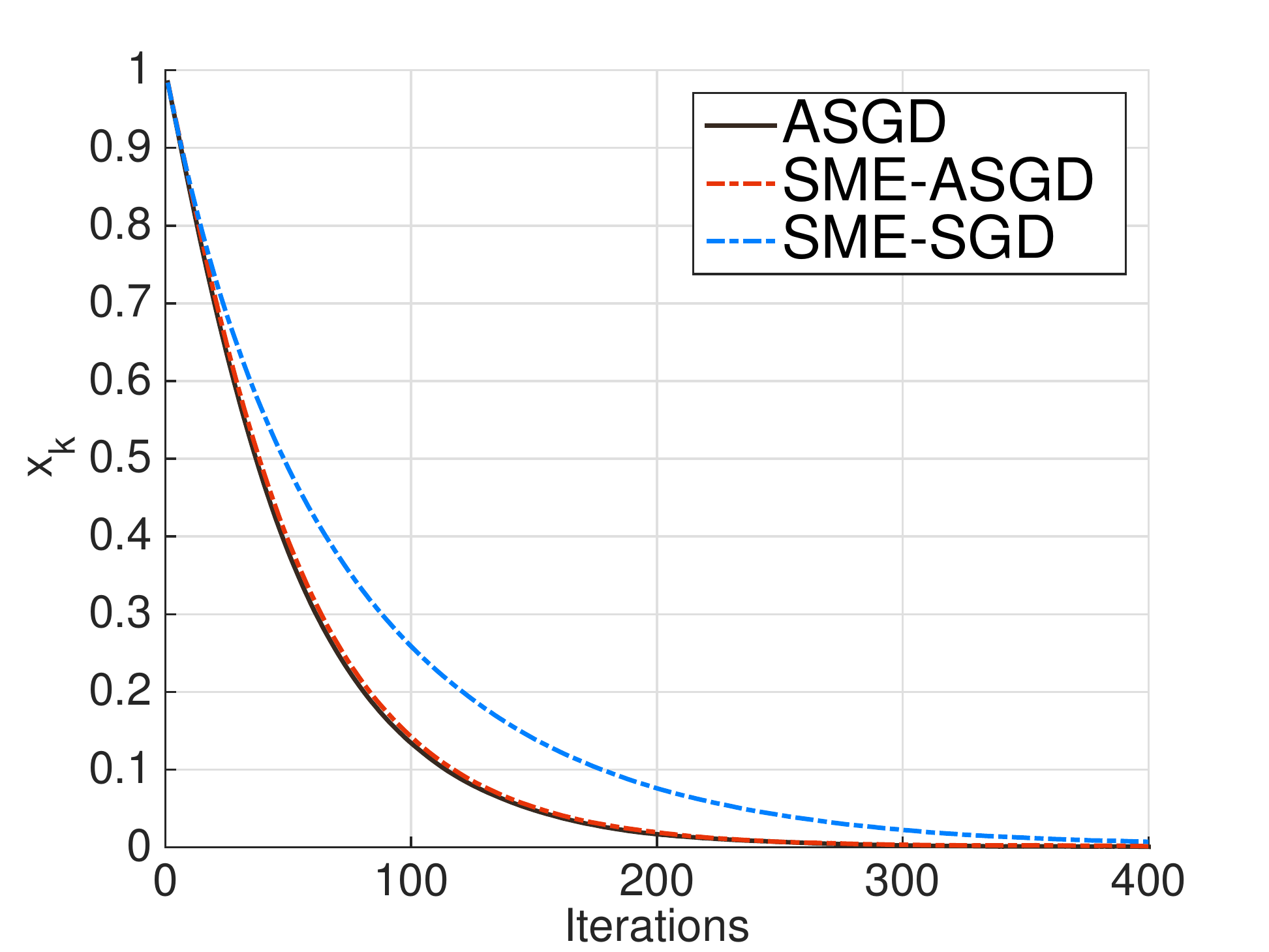} 
    \end{minipage}\qquad
    \begin{minipage}{0.4\textwidth}
        \centering
        \includegraphics[width=0.95\textwidth]{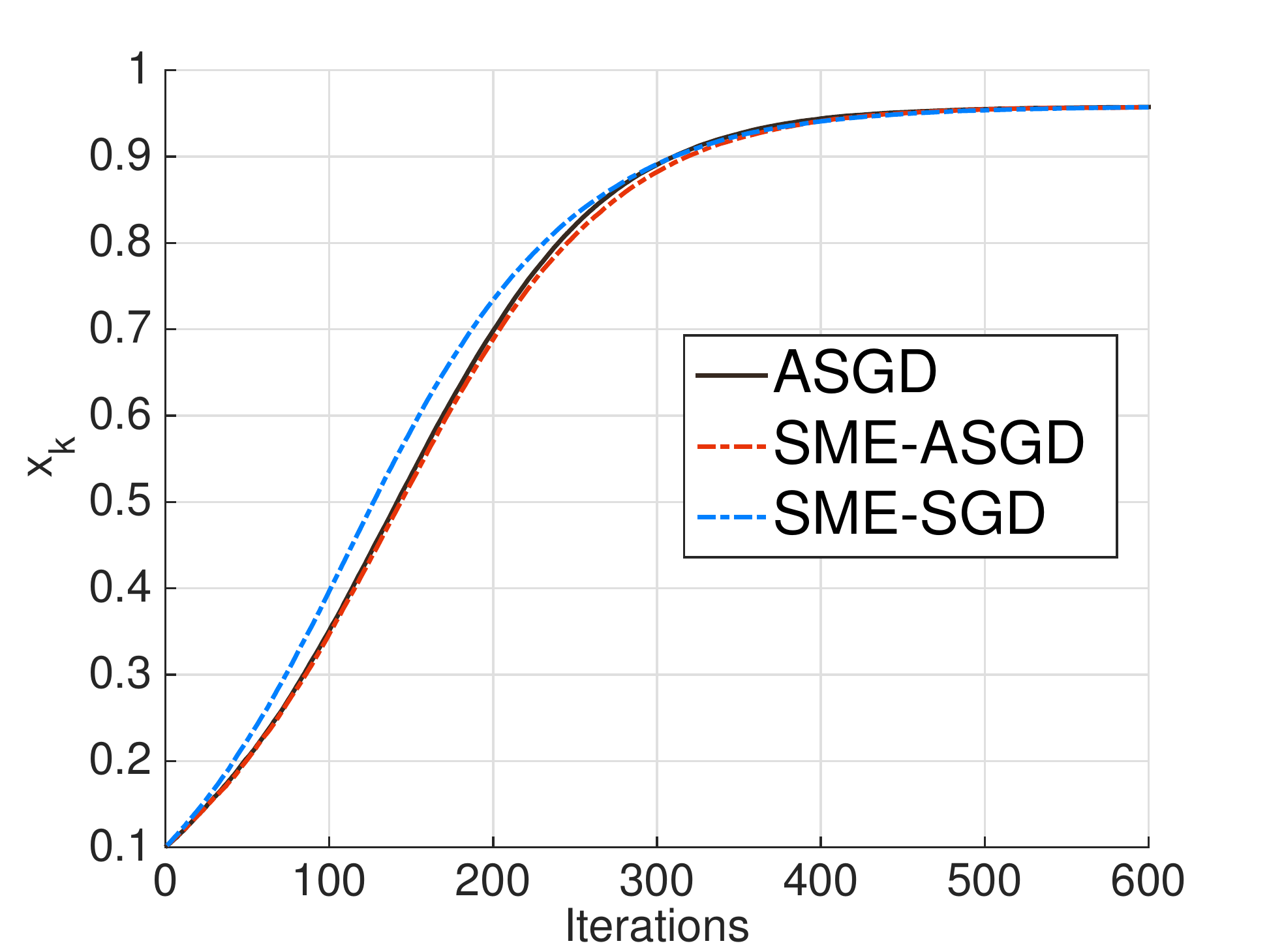} 
    \end{minipage}
    \begin{minipage}[a]{0.4\textwidth}
        \centering
        \includegraphics[width=0.95\textwidth]{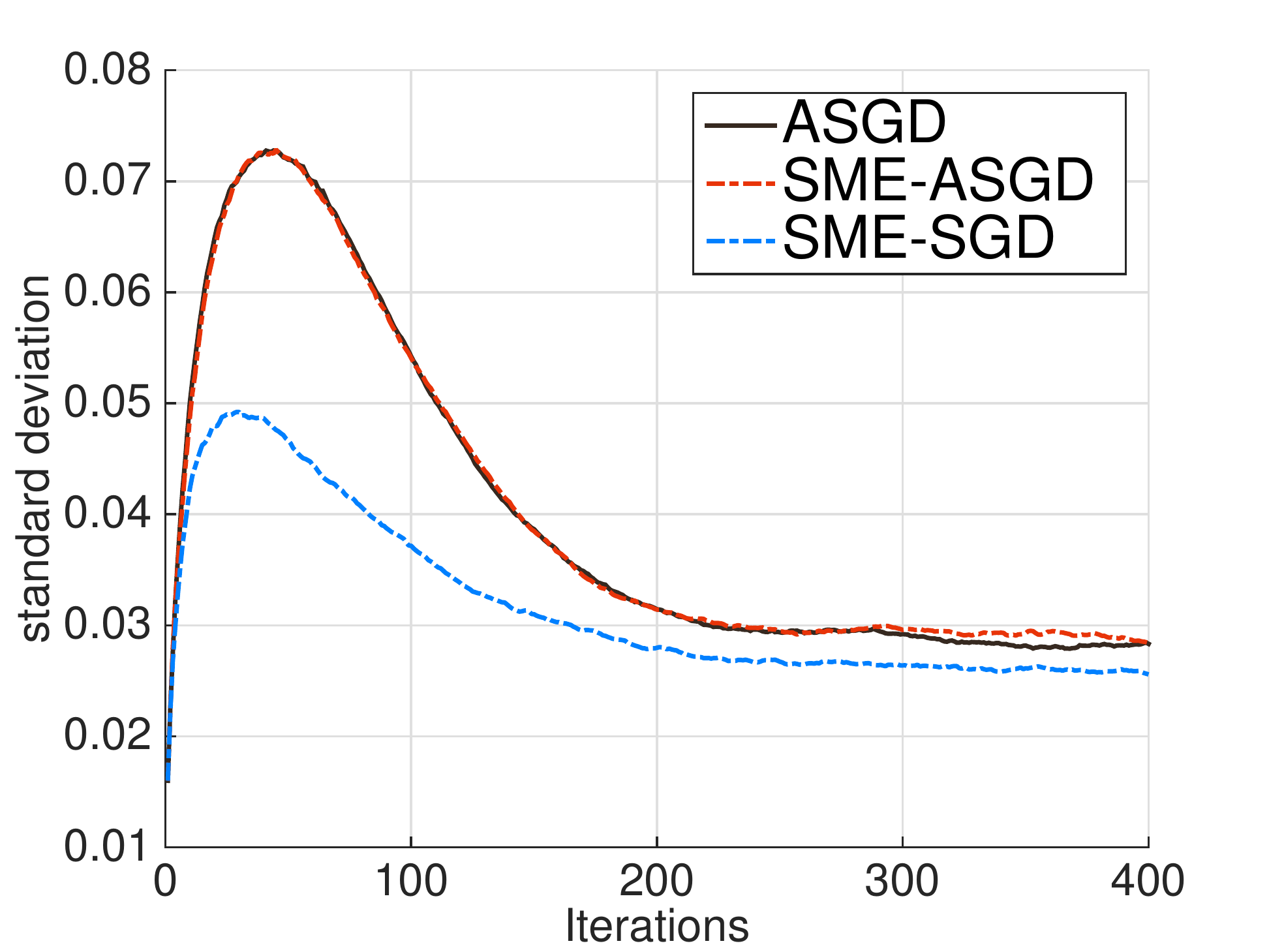} 
    \end{minipage}\qquad
        \begin{minipage}{0.4\textwidth}
        \centering
        \includegraphics[width=0.95\textwidth]{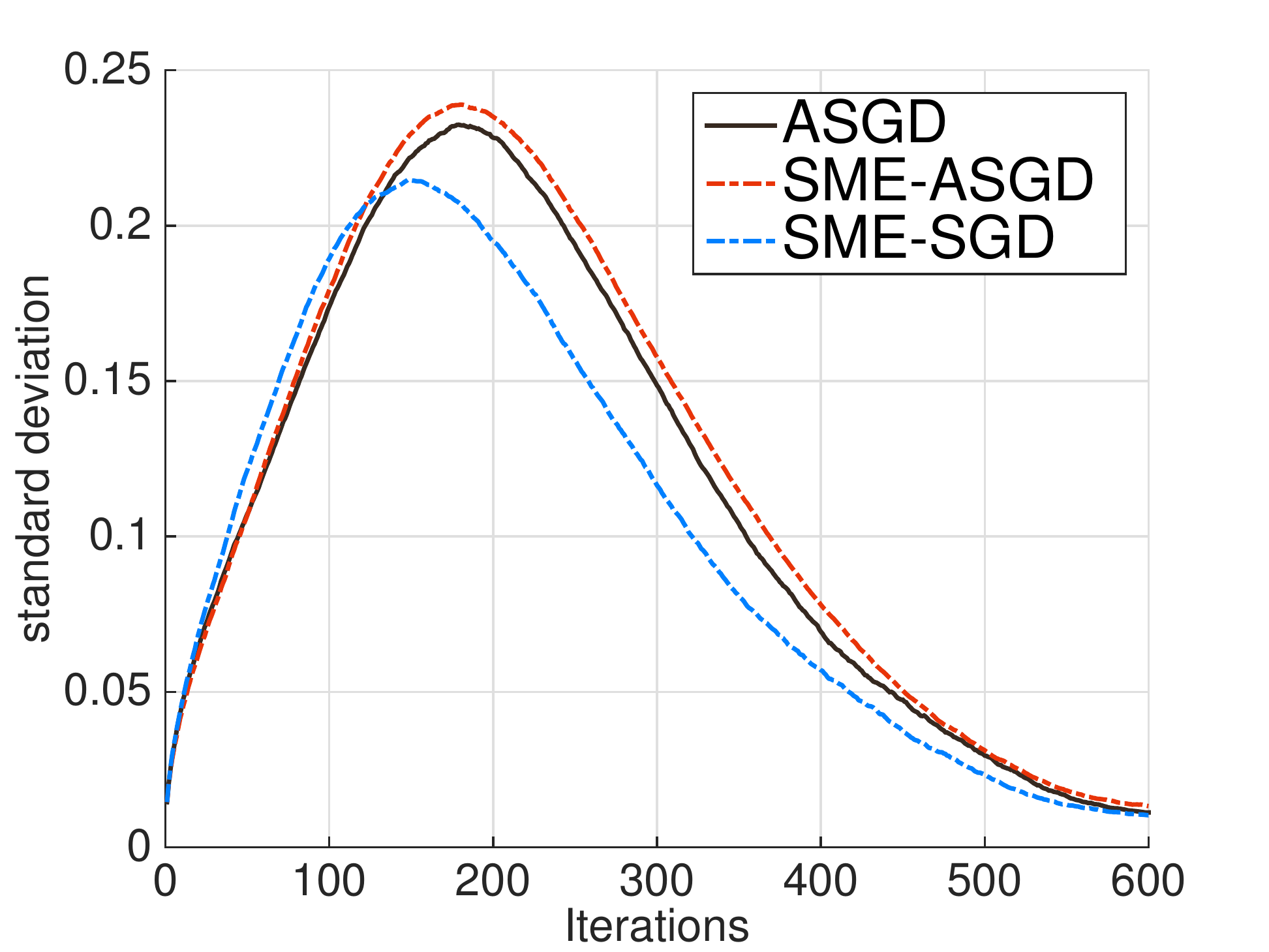} 
      \end{minipage}
    \caption{ (Left) Apply the SME-ASGD to minimize the convex function $f(x) = x^4+ 6x^2$ with
      two components $f_1(x) = (x-1)^4-1$, and $f_2(x) = (x+1)^4-1$. Notice that the gradients are
      Lipschitz locally. Here we choose $x_0 =1$, and a smaller step size $\eta = 1e-3$. (Right)
      Apply the SME-ASGD to minimize the double well potential $f(x) = 1 - e^{-(x-1)^2} -
      e^{-(x+1)^2}$. Here $f_1 = 1 - 2e^{-(x-1)^2}, f_2 = 1 - 2e^{-(x+1)^2}$ and both have Lipschitz
      gradients. We choose $ \eta = 1e-2, x_0 = 0.1$.
      Note that $\argmin f(x)\approx\pm0.9575$. In our case, due to the initial data $x_0$,
      $90.34\%$ of ASGD path samples converge to $0.9575$, while $90.50\%$ of SME-ASGD and $88.54\%$
      of SME-SGD converge to the same minimizer. For both columns of numerical tests, we choose $\mu
      = 0.95$.}
    \label{fig:nonlinear}
\end{figure}

Here, we provide some numerical evidences for Theorem \ref{thm} with various loss functions
$f$. The results are shown in Figures~\ref{fig:linear} (for linear forcing) and \ref{fig:nonlinear}
(for general forcing).  For each example, through averaging over $5000$ samples, we compare the
results of ASGD with the predictions from both SME-ASGD \eqref{nonlinearASGD} and the 2nd-order weak
convergent SME-SGD proposed in Li et al.'s paper \cite{li2015stochastic}
\begin{align}\label{LiSME}
  dX_t = -\nabla (f(X_t) + \frac{\eta}{4}|\nabla f(X_t)|^2) dt +  (\eta \Sigma(X_t))^{1/2} dB_t.
\end{align}
When $\mu$ is close to $0$ (i.e., the expected delay is short), 
SME-SGD \eqref{LiSME} serves as a good approximation to ASGD as expected. However, when $\mu$ is large, 
Figures \ref{fig:linear} and \ref{fig:nonlinear} demonstrate that it is no longer the case: 
As $\mu$ gets closer to $1$, the trajectories obtained from SME-SGD are way off, 
whereas our proposed SME-ASGD model demonstrate accurate path approximations for both the first 
and the second moments.

\begin{figure}[h!]
    \centering
    \begin{minipage}{0.32\textwidth}
        \centering
        \includegraphics[width=0.95\textwidth]{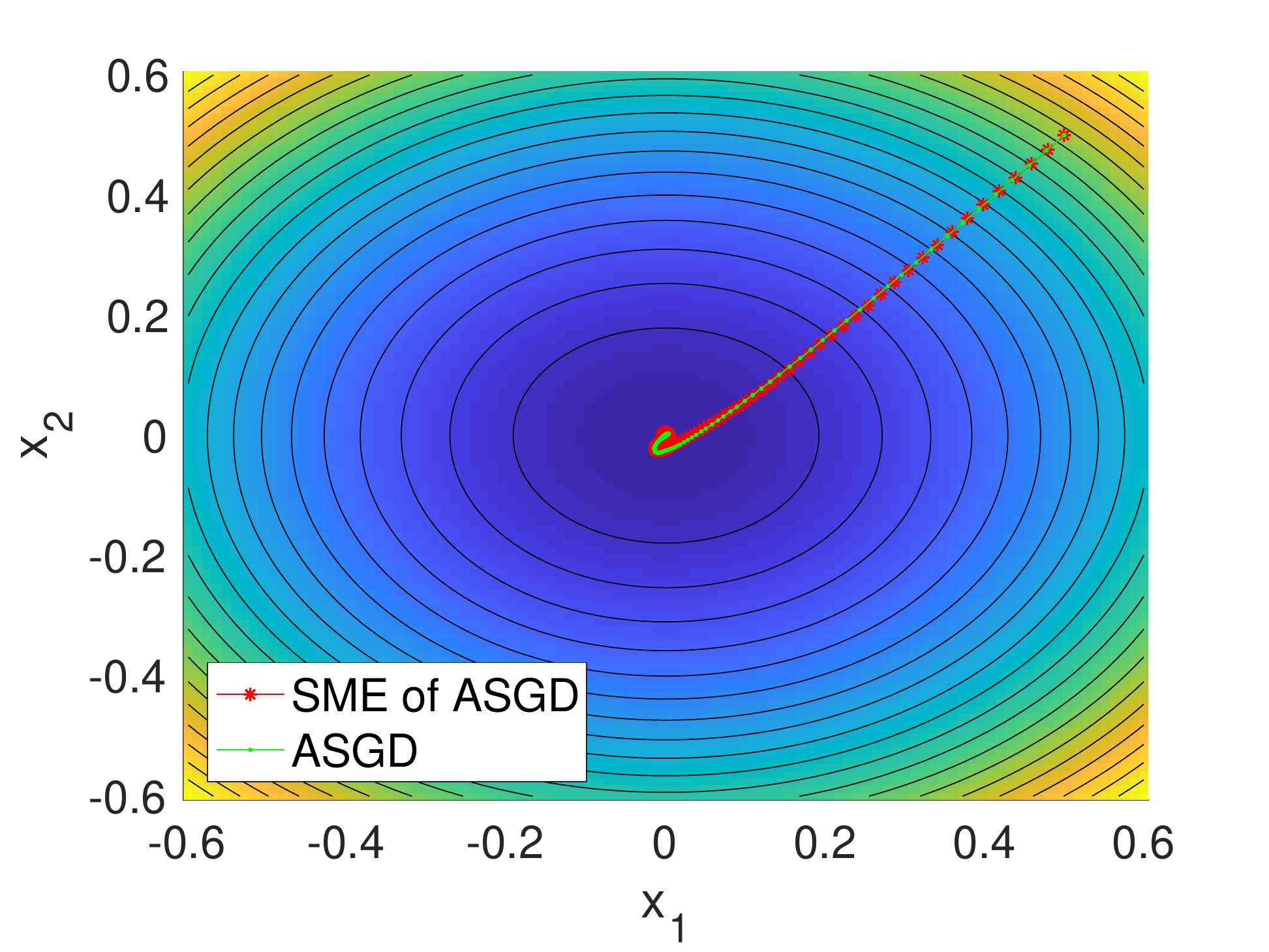} 
    \end{minipage}\hfill
    \begin{minipage}{0.32\textwidth}
        \centering
        \includegraphics[width=0.95\textwidth]{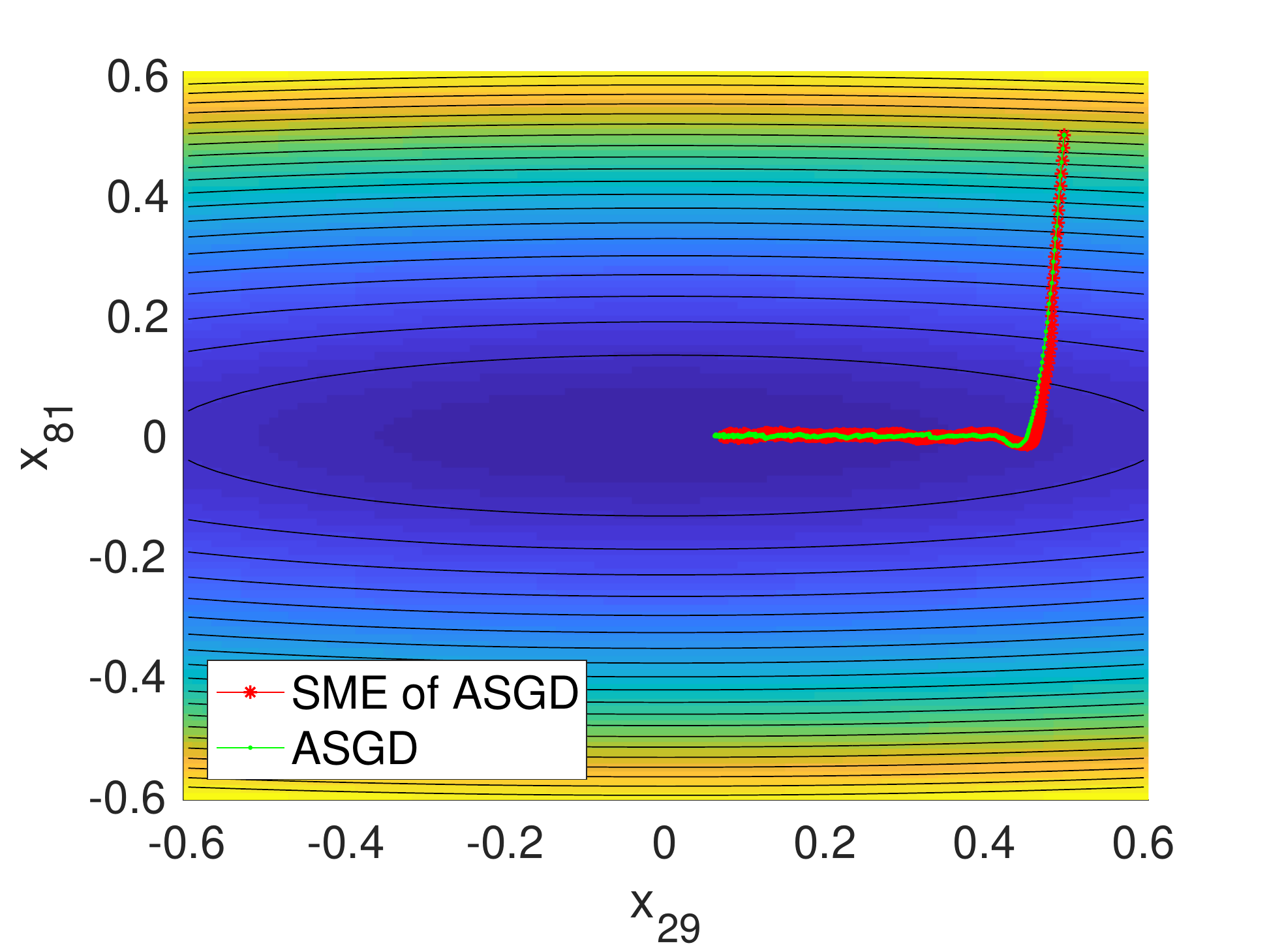} 
    \end{minipage}
        \begin{minipage}{0.32\textwidth}
        \centering
        \includegraphics[width=0.95\textwidth]{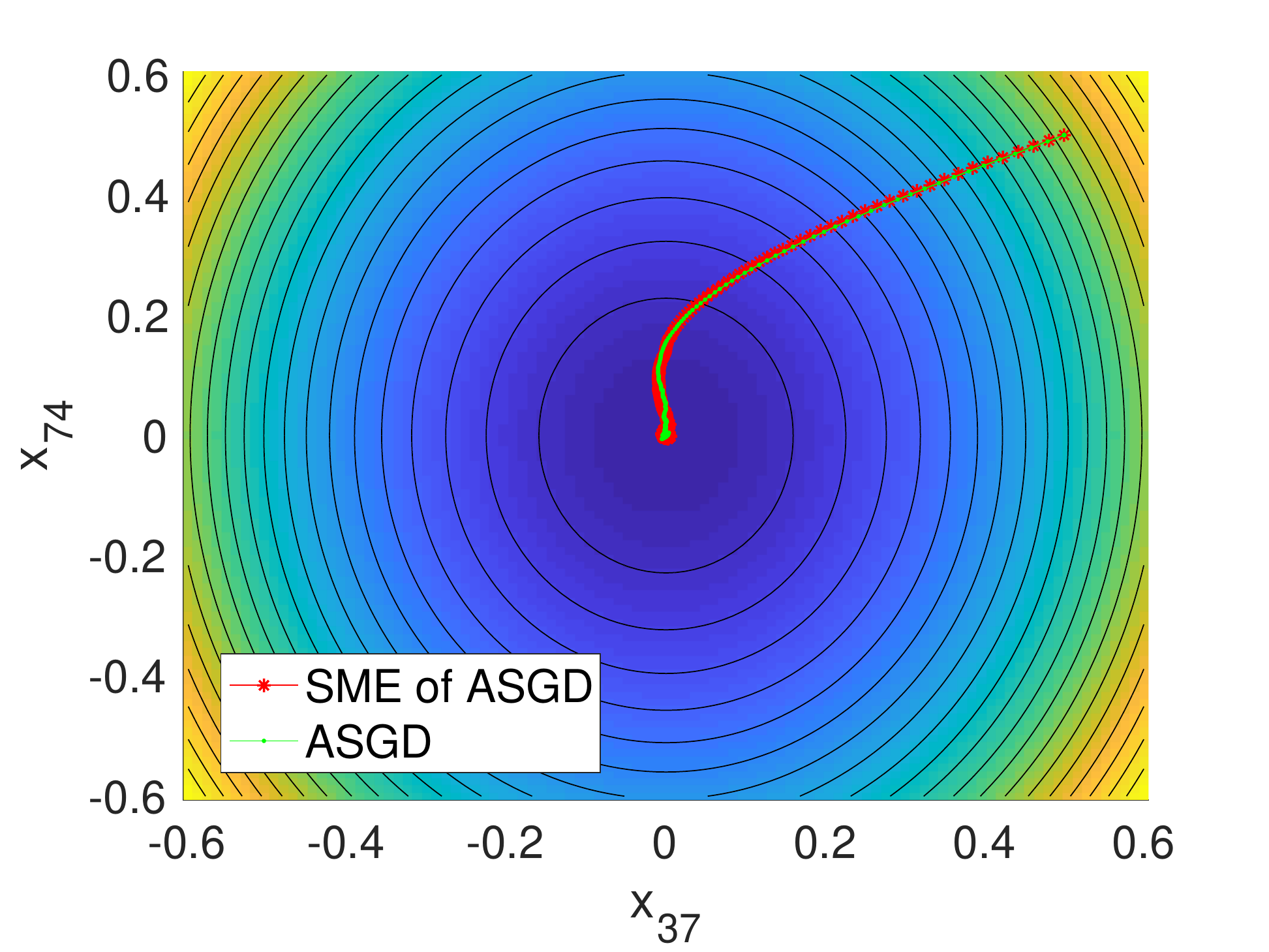} 
    \end{minipage}
    \caption{Apply the SME-ASGD to minimize the quadratic function $f(x) = \sum_{i=1}^{100} c_i x_{i}^2/2, ~ x\in \mathbb{R}^{100}$ with $\mu=0.90$ and two components $f_1(x) = \sum_{i=1}^{100} c_ix_i^2/2-x$, and $f_2(x) = \sum_{i=1}^{100} c_ix_i^2/2+x$. The initial condition $x_0 = (0.5,0.5,\cdots, 0.5)\in \mathbb{R}^{100}$ and the step size is $\eta= 1e-2$. The plots are done after $1000$ iterations. The corresponding coefficients in the plots are $c_1 = 4.2593, c_2= 4.9013, c_{29} = 0.1980, c_{81} = 4.3968, c_{37} = 3.9978, c_{74}=1.9527$.  }
    \label{fig:nd1}
    \begin{minipage}[a]{0.32\textwidth}
        \centering
        \includegraphics[width=0.95\textwidth]{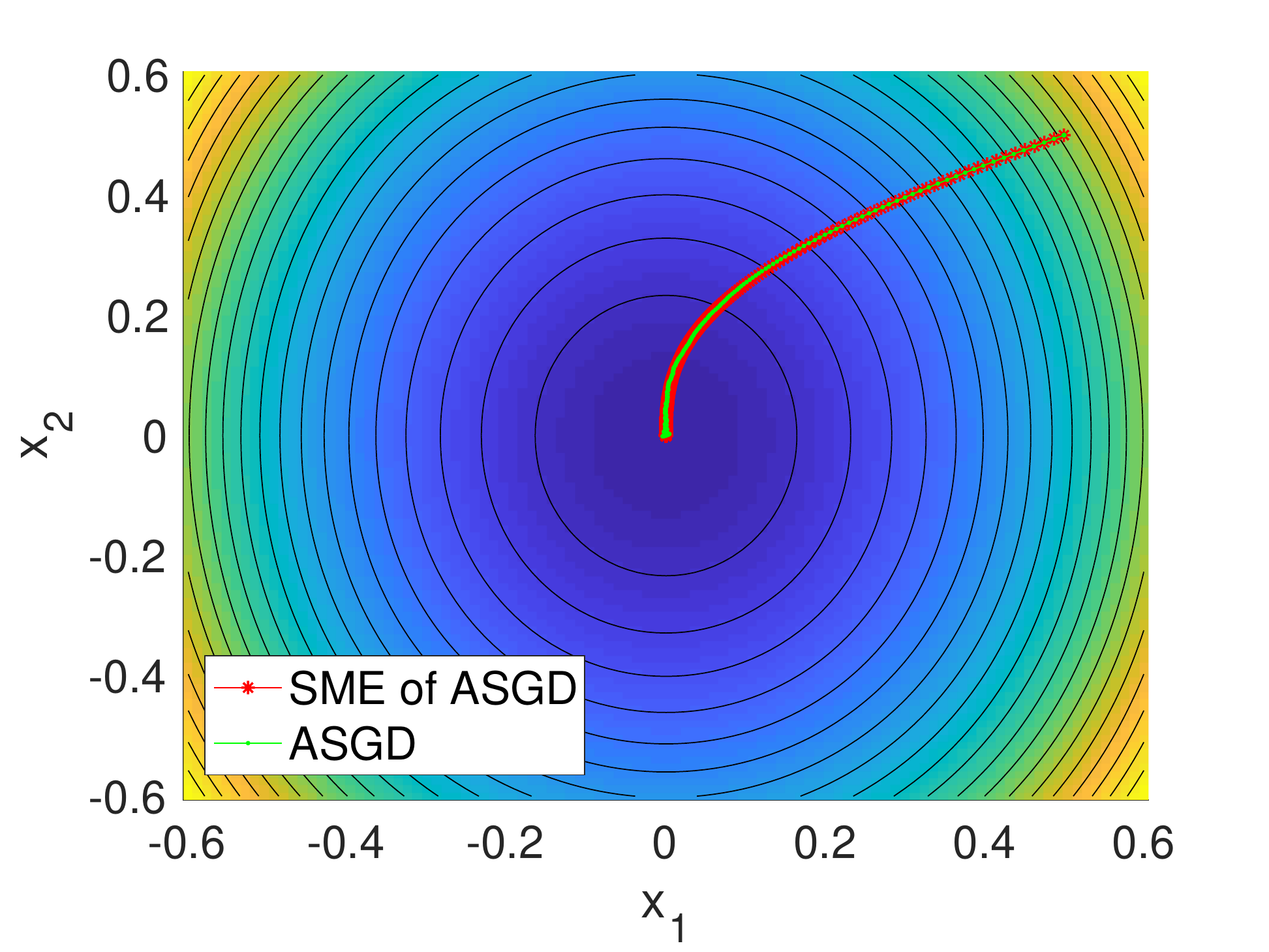} 

    \end{minipage}\hfill
        \begin{minipage}{0.32\textwidth}
        \centering
        \includegraphics[width=0.95\textwidth]{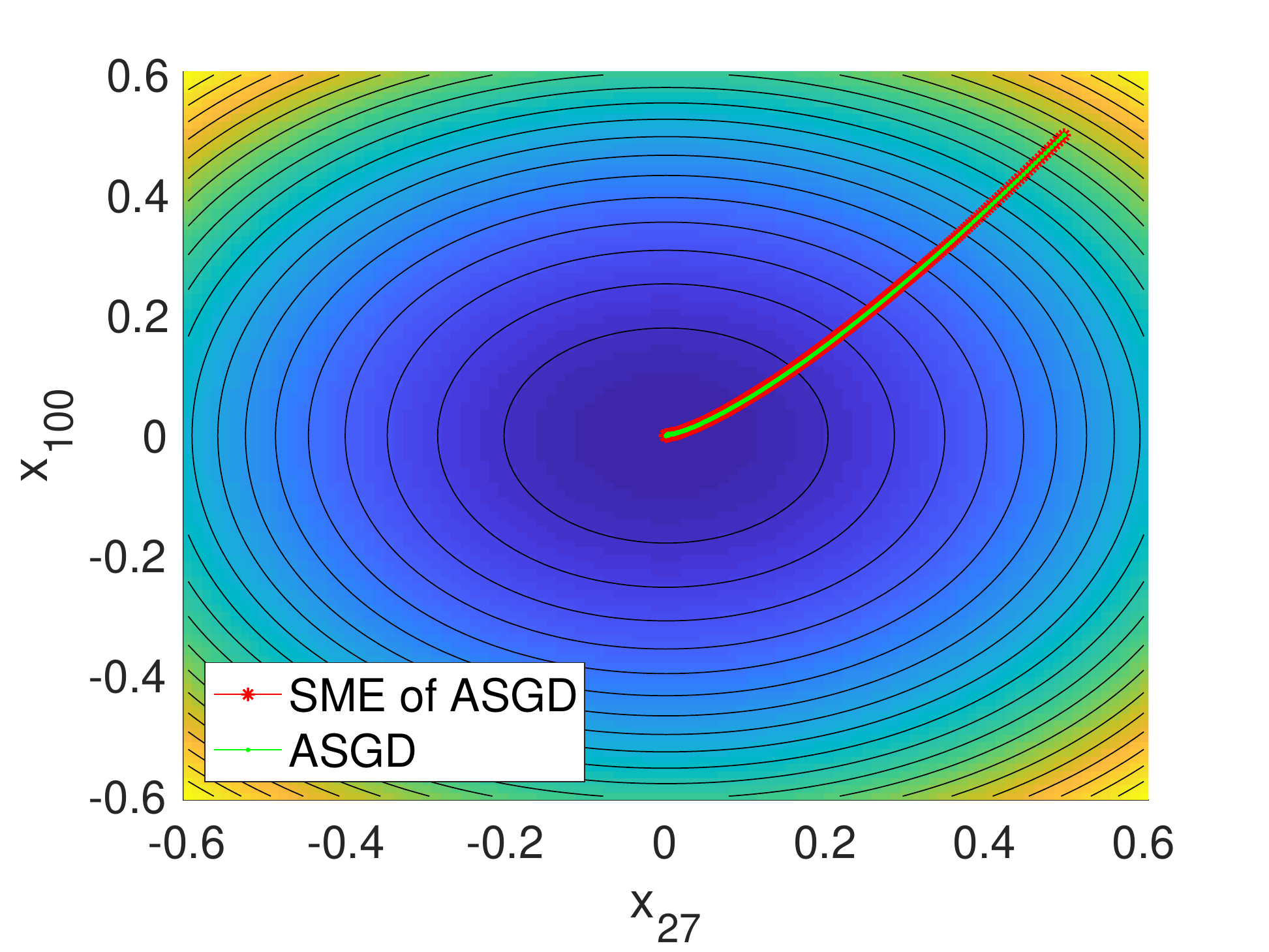} 

    \end{minipage}
    \begin{minipage}{0.32\textwidth}
        \centering
        \includegraphics[width=0.95\textwidth]{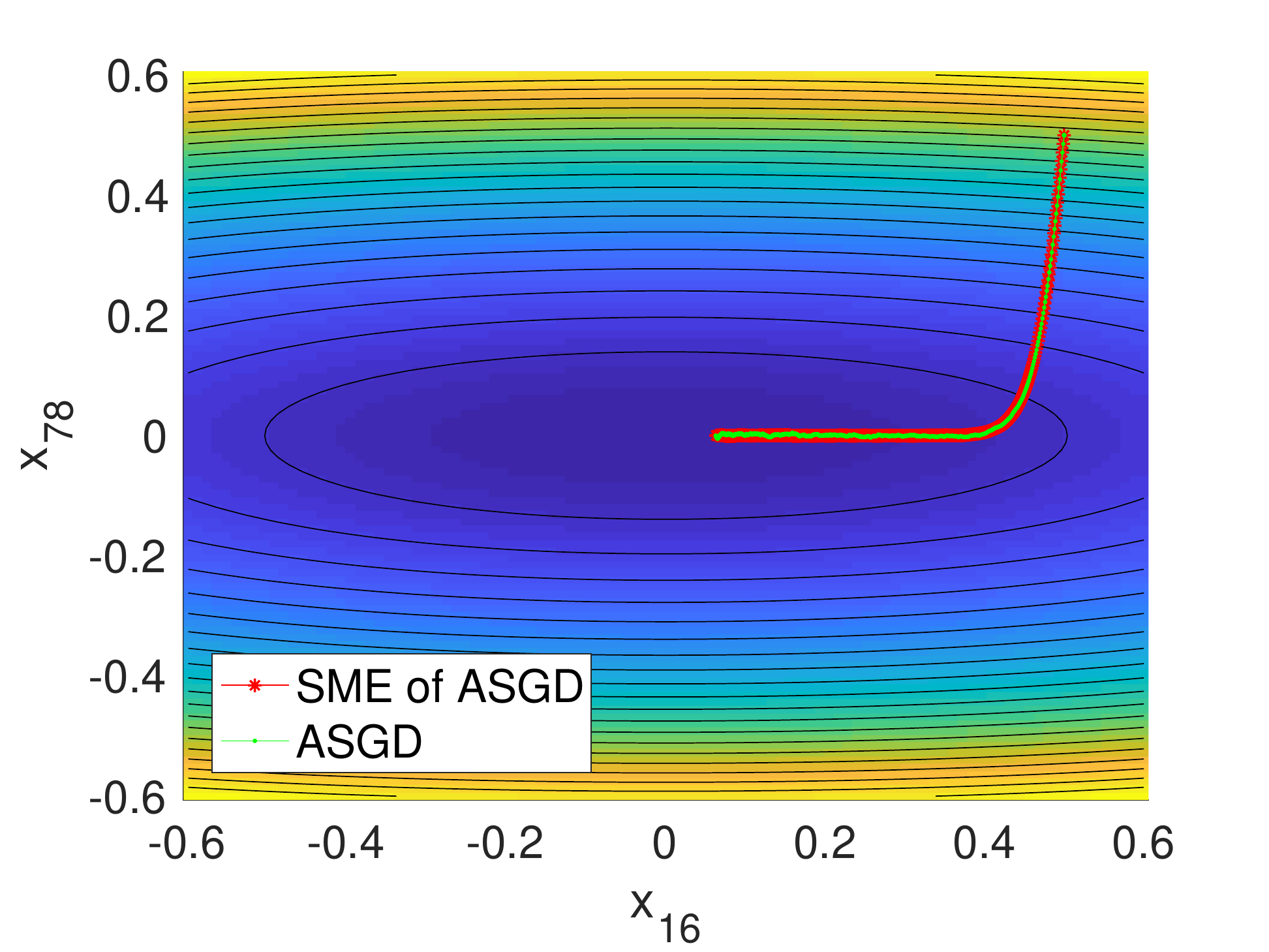} 

    \end{minipage}
     \caption{ Apply the SME-ASGD to minimize the convex function $f(x) = \sum_{i=1}^{100} c_i (x_{i}^4+6x_{i}^2)/2, ~ x\in \mathbb{R}^{100}$ with $\mu=0.90$ and two components $f_1(x) = \sum_{i=1}^{100} c_i(x_i-1)^4/2-1$, and $f_2(x) = \sum_{i=1}^{100} c_i(x_i+1)^4/2+1$. The initial condition $x_0 = (0.5,0.5,\cdots, 0.5)\in \mathbb{R}^{100}$ and the step size is $\eta= 1e-3$. The plots are done after $1000$ iterations. The corresponding coefficients in the plots are $c_1 = 3.9212, c_2= 1.9370, c_{27} = 1.2093, c_{100} = 1.5661, c_{16} = 0.3353, c_{78}=4.5502$. }
     \label{fig:nd2}
\end{figure}

A few remarks regarding the numerical results are in order here. (i) In Figure \ref{fig:linear},
the path oscillations happen to both ASGD and SME-ASGD due to a longer expected delay, but not to
SME-SGD, even though we include staleness when computing $\Sigma(X_t)$ by the convariance matrix formula for both models. That is
because our SME-ASGD model contains $\mu$ in the forcing term, while the forcing term in SME-SGD is
$\mu$-independent.
(ii) The convex function $f(x)=x^4+6x^2$ (with gradient $\nabla f(x) = 4x^3+12x$) in Figure
\ref{fig:nonlinear} does not satisfy the general Ito conditions; however, by having good initial
data and choosing smaller time step sizes, we can still obtain the minimizer without blowing
up.
(iii) For the non-convex example (the double-well function in Figure \ref{fig:nonlinear}), the
SME-ASGD model gives a better prediction about which minimizer that a trajectory with
given initial data will fall into: The percentage of path samples that converge to a local minimum 
in SME-ASGD is very close to that of the ASGD case. 
(iv) For all cases, SME-SGD underestimates the variance because the variance from the delayed reads 
is not taken into account by SME-SGD. (v) In higher dimensions, unlike the Monte Carlo sampling driven by Langevin dynamics that has the curse of dimensionality issue, our numerical simulations for both linear and nonlinear gradidents have good approximation regardless of the dimensionality as the Figures \ref{fig:nd1} and \ref{fig:nd2} show. Here, we assign the coefficients $c_i$ uniformly randomly in $[0,5]$. We make plots by arbitrarily choosing any two dimension as projected subspace. Although after $1000$ time steps, some projected subspace have convergence and and some (with significant coefficient differences) do not yet, we can see that the trajectories from the algorithm and modified equation are close.

\section{Optimal mini-batch size of ASGD}
With much better understanding of dynamics of the ASGD algorithm using SME-ASGD, we are able to tune
multiple hyper-parameters of ASGD using the predictions obtained from applying the stochastic optimal
control theory to SME-ASGD. Here we demonstrate one such application: the optimal time-dependent
mini-batch size for ASGD. By denoting the time-dependent batch size as $1+u_k$ with $ u_k\geq 0$,
one can write the iteration as
\begin{align}\label{batch}
  x_{k+1} = x_k -\eta \frac{1}{1+u_k} \sum_{j=1}^{1+u_k} \nabla f_{\gamma_j}(x_{k-\tau_k}).
\end{align}
We argue that it is reasonable to assume that the choice of mini-batch size is independent from $\gamma_j$ and the 
staleness $\tau_k$. This is because, even though changing the batch size will simultaneously change the "clocks"
of all the processors, the staleness would not be changed as all the processors are impacted
equally. Following the argument given in Section 2, we can derive a corresponding SME
\begin{align}\label{batchsme}
  \nonumber
  &dY_t = -\nabla f(X_t) dt -\sqrt{\frac{1-\mu}{\eta}} Y_t dt \\
  &dX_t = Y_t dt + \frac{\sigma(t) \eta^{3/4}}{(1+u(t))^{1/2}(1-\mu)^{1/4}} dB_t.
\end{align}
The derivation here is not much different from the one of SME-ASGD \eqref{nonlinearASGD}, 
except for identifying the right coefficient in front of the the noise term $dB_t$. 
The correct coefficient (denoted by $c$ in the discussion below) is constrained by the 
following constraints on the variance
\begin{align*}
  \mathbb{E} & \left\{\frac{\eta^2}{(1+u_k)^2} \left( \sum_{j=1}^{1+u_k} (-\frac{y_k}{\alpha} - \nabla
  f_{\gamma_j}(x_{k-\tau_k}) \right) \left( \sum_{j=1}^{1+u_k} (-\frac{y_k}{\alpha} - \nabla
  f_{\gamma_j}(x_{k-\tau_k}) \right)^T\right\} \\ & = \frac{\eta^2}{(1+u_k)^2}\sum_{j=1}^{1+u_k}
  \mathbb{E}\left\{\left(-\frac{y_k}{\alpha} - \nabla f_{\gamma_j}(x_{k-\tau_k})\right) \left(-\frac{y_k}{\alpha} -
  \nabla f_{\gamma_j}(x_{k-\tau_k})\right)^T \right\} = \frac{\eta^2}{1+u_k} \Sigma_k \sim c^2 \Delta t,
\end{align*}
where the cross terms vanish under the expectation. Plugging in $\Delta t = \sqrt{\eta(1-\mu)}$ shows that 
the coefficient for the noise is
\begin{align*}
  c = \frac{\sigma(t) \eta^{3/4}}{(1+u(t))^{1/2}(1-\mu)^{1/4}}
\end{align*}
as shown in \eqref{batchsme}.

We would like to explore the dynamics of SME to find the dominating eigenvalue for later use.
To simplify the discussion, let us consider for example the quadratic loss objective $f(x) =
x^2$. By applying the Ito's formula to this SME, one obtains the following evoluation system for the
second moments
\begin{align}\label{secmomen}
  \frac{d}{dt}\begin{bmatrix} \mathbb{E}(X_t^2)\\ \mathbb{E}(Y_t^2) \\ \mathbb{E}(X_t Y_t) \end{bmatrix} &= - \begin{bmatrix} 0 & 0 & -2\\ 0 & 2\sqrt{(1-\mu)/\eta} & 4 \\ 2 & -1 & \sqrt{(1-\mu)/\eta}  \end{bmatrix}\begin{bmatrix} \mathbb{E}(X_t^2)\\ \mathbb{E}(Y_t^2) \\ \mathbb{E}(X_t Y_t) \end{bmatrix}   + \begin{bmatrix}  \frac{\Sigma(t)\eta^{3/2}}{(1+u(t))(1-\mu)^{1/2}} \\ 0 \\ 0\end{bmatrix}.  
\end{align}
A similar derivation is shown in Appendix B, and we just replace all $\Sigma$ by $\Sigma/(1+u(t))$ 
in the mini-batching case. Here, we make a simplifying but practical assumption that $u(t)$ varies 
slowly. Now by freezing $u(t)$ to a constant $u$, \eqref{secmomen} is a linear system with constant coefficients, 
its asymptotic behavior is determined by the eigenvalue of the
coefficient matrix. An easy calculation shows that the eigenvalue with
largest real part is given by
$\lambda =-\sqrt{\nicefrac{(1-\mu)}{\eta}} + \sqrt{\nicefrac{(1-\mu -
    8\eta)}{\eta}}$
with a negative real part and therefore the second moment of $X_t$ decays exponentially.  Moreover, \eqref{secmomen} provides us with the stationary solution for $X^2$
\begin{align}
  z_{\infty} := \mathbb{E}(X_{\infty}^2) = \frac{\Sigma \eta}{2(1+u(t))} \Bigl( \frac{\eta}{1-\mu} + \frac{1}{2}\Bigr).
\end{align} 
For a slowly varying $u(t)$, $z_{\infty}=z_{\infty}(u(t))$ is a function of $u(t)$. 
Based on this simplication, rather than applying the optimal control subject to the full second moment equation, 
we shall work with a simpler evolution equation that asymptotically approximates the dynamics (imposed as a
constraint). More specifically, we pose the following optimal control problem for the time-dependent
mini-batch size
\begin{align}\label{opt}
  & \min_{u\in \mathcal{A}} 
    \bigg\{ z(T) + \frac{\gamma}{\eta}\int_0^T u(s) ds \bigg\}    \,\,\,\,\, \text{ subject to }\\
    &\nonumber \frac{d}{dt}z(t) = \mathrm{Re}(\lambda ) (z(t) - z_{\infty}(u(t))) 
  \qquad  \text{ with } z(0) = x_0^2,
\end{align}
where $z(t)$ models $\mathbb{E}(X_t^2)$ -- the quantity to minimize,
$\mathcal{A}=\{u(t)\geq 0\}$ is an admissible control set as the
mini-batch size is greater than $1$, and $\gamma>0$ is a constant
measuring the unit cost for introducing extra gradient samples
throughout the time. Below we show how to solve the optimal control problem \eqref{opt}. The value function can be
defined as
\begin{align}
  V(z,t) = \min_{u\in \mathcal{A}}\bigg\{ z(T) + \frac{\gamma}{\eta}\int_t^T u(s) ds \; \bigg|\;
  \frac{d}{dt}z(t)=F(u(t),z(t)), z(t) = z \bigg\},
\end{align}
where $F(u(t),z(t))=\text{Re}(\lambda ) (z(t) -z_{\infty}(u(t))) = \text{Re} (\lambda)\bigl(z(t)
-\frac{\Sigma \eta}{2(1+u(t))} ( \frac{\eta}{1-\mu} + \frac{1}{2})\bigr)$. The corresponding Hamilton-Jacobi-Bellman equation
is
\begin{align}
  &V_t +\min_{u\in \mathcal{A}}\bigg\{ F(u,z) V_z + \frac{\gamma}{\eta} u \bigg\} = 0\\
  & \nonumber \text{with } V(0, t) =0, V(z,T) = z.
\end{align}
Since $\min_{u\in \mathcal{A}}\bigg\{ F(u,z) V_z + \frac{\gamma}{\eta} u \bigg\} = \min_{u\in
  \mathcal{A}}\bigg\{ \frac{-V_z \text{Re}(\lambda ) \Sigma\eta}{2(1+u)} \big(
\frac{\eta}{1-\mu} + \frac{1}{2}\big)+ \frac{\gamma}{\eta} u \bigg\}$, $V_z \geq 0$, and
$\text{Re}(\lambda )<0$, the minimum could be obtained by solving the following equation
\[
\frac{V_z \text{Re}(\lambda ) \Sigma\eta}{2(1+u)^2} \big( \frac{\eta}{1-\mu} + \frac{1}{2}\big)+ \frac{\gamma}{\eta} = 0
\]
with the derivative of the value function $V_z$ to be determined
later. Therefore the optimal batch size $u^*$ as a function of $V_z$ is
\begin{align}\label{utemp}
u^*(V_z)=
\begin{cases}
\sqrt{\frac{-V_z \text{Re}(\lambda ) \Sigma\eta^2}{2\gamma}\big( \frac{\eta}{1-\mu} + \frac{1}{2}\big)} -1 & \text{if } \frac{-V_z \text{Re}(\lambda ) \Sigma\eta^2}{2\gamma} \big( \frac{\eta}{1-\mu} + \frac{1}{2}\big)> 1, \\
0 & \text{otherwise}.
\end{cases}
\end{align}
The next step is to solve $V$ to get an explicit formula for $u^*$. Placing $u^*(V_z)$ back into the
minimization bracket, we obtain
\begin{align}
  \min_{u\in \mathcal{A}}\bigg\{ F(u,z) V_z + \frac{\gamma}{\eta} u \bigg\} =
  \begin{cases}
    \text{Re}(\lambda ) z V_z - \frac{\gamma}{\eta}& \text{if } \frac{-V_z \text{Re}(\lambda) \Sigma\eta^2}{2\gamma} \big( \frac{\eta}{1-\mu} + \frac{1}{2}\big)> 1, \\
    \text{Re}(\lambda ) \big(z-\frac{\Sigma \eta}{2}\big( \frac{\eta}{1-\mu} + \frac{1}{2}\big)\big) V_z & \text{otherwise}.
  \end{cases}
\end{align}
This gives the Hamilton-Jacobi equation and we can solve it by using the method of
characteristics. Letting $\gamma^* = -\frac{\text{Re}(\lambda ) \Sigma\eta^2}{2}\big(
\frac{\eta}{1-\mu} + \frac{1}{2}\big)$ for notation convenience, we obtain the solution for $V$
\begin{align}
V(z,t)=
\begin{cases}
  \frac{\Sigma\eta}{2}\big( \frac{\eta}{1-\mu} + \frac{1}{2}\big) + \big(z-\frac{\Sigma\eta}{2}\big( \frac{\eta}{1-\mu} + \frac{1}{2}\big) \big)e^{\text{Re}(\lambda ) (T-t)}& \text{if } \gamma> \gamma^* \\
  \big(z-\frac{\Sigma\eta}{2}\big( \frac{\eta}{1-\mu} + \frac{1}{2}\big) \big)e^{\text{Re}(\lambda ) (T-t)} -\frac{\gamma}{\eta}(t^*+\frac{1}{\text{Re}(\lambda )})& \text{if } \gamma\leq\gamma^*, 0\leq t\leq T-t^*\\
  ze^{\text{Re}(\lambda) (T-t)} - \frac{\gamma}{\eta}(T-t)& \text{if } \gamma\leq \gamma^*, T-t^*<t\leq T,
\end{cases}
\end{align}
where  $  t^* = \frac{1}{\text{Re}(\lambda )}\log(\frac{\gamma}{\gamma^*})$. For all cases, $V_z = e^{\text{Re}(\lambda ) (T-t)}$. With this inserted back into
\eqref{utemp}, we conclude that

\begin{align}\label{optsoln}
  u^*(t)=
  \begin{cases}
    0& \text{if } \gamma> \gamma^* \text{ or } 0\leq t\leq T-t^* \\
    \sqrt{\frac{\gamma^*}{\gamma} }e^{\text{Re}(\lambda )(T-t)/2} - 1& \text{if } \gamma\leq \gamma^*, T-t^*<t\leq T.
  \end{cases}
\end{align}

\begin{figure}[h]
  \centering \includegraphics[width=0.5\textwidth]{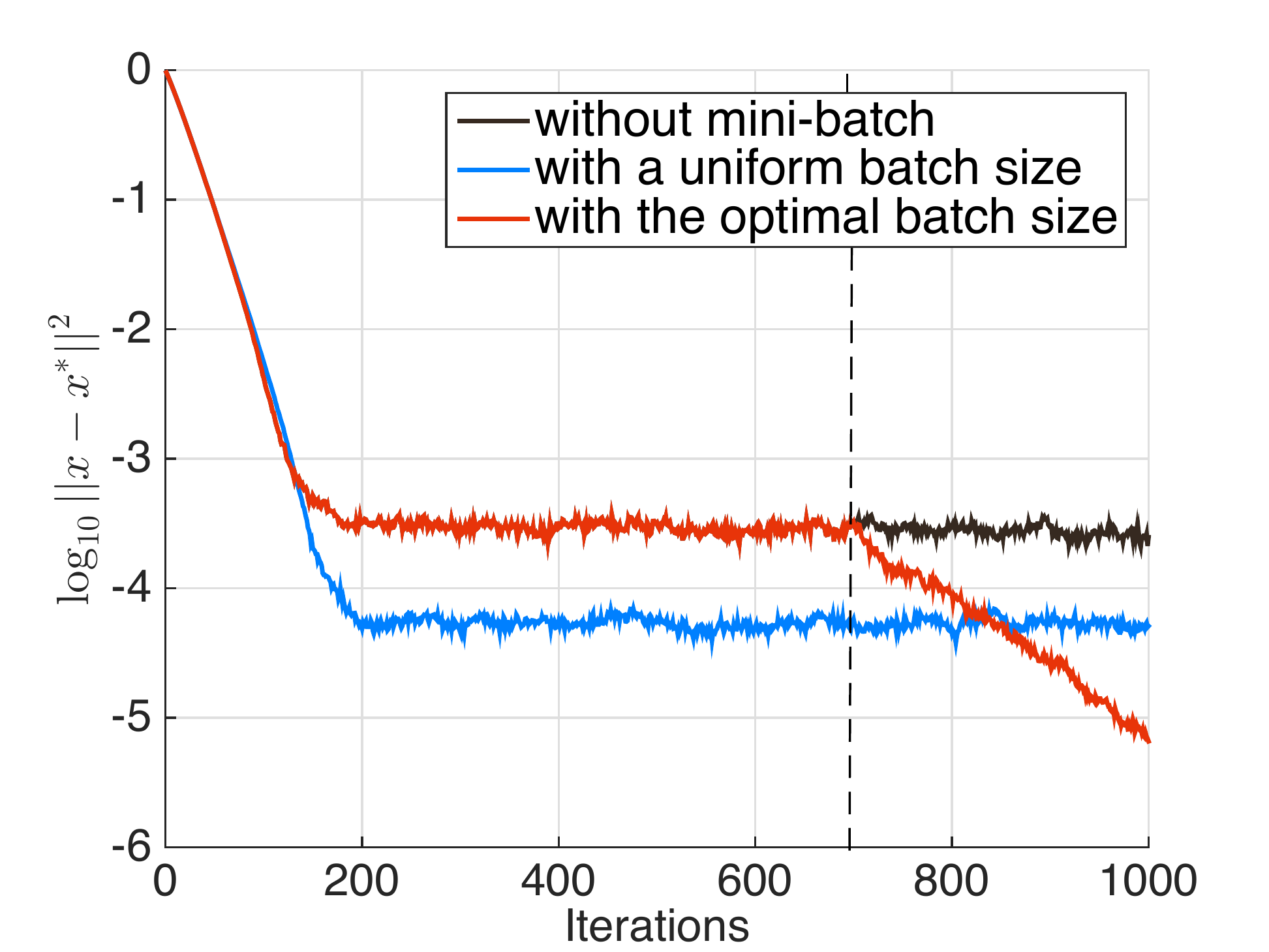}
  \caption[Controlled mini-batch sizes] {\small A comparison of
    performance in terms of $l^2$ error. We apply mini-batching over
    $n=100$ components $f_i(x) = \frac{1}{2}(x-c_i)^2, c_i =-1/2+i/(2n)$. Here we choose
    the step size $\eta = 0.02$ and the initial data $x_0 = 1$. The
    batch size for the uniform mini-batching case is $5$. For the
    optimal mini-batching strategy, the transition happens at
    $k = (T-t^*)/\eta \approx 699$, and the optimized batch size at
    time $T$ is $42$. In practice, we can apply a more aggressive
    mini-batching strategy by starting to increase the batch size
    earlier in the flat region, and it will result in a larger batch
    size at $T$.  }
  \vspace{-1em}
  \label{fig:opt}
\end{figure}

In particular, \eqref{optsoln} tells that we should use a small
mini-batch size (even size $1$) during the early time (for
$k \leq k^*=(T-t^*)/\eta$), since during this period the gradient flow
dominates the dynamics. After the transition time $k^*$ at which the noise
starts to dominate, one shall apply mini-batch with size exponentially
increasing in $k$ to reduce the variance. Figure \ref{fig:opt}
demonstrates that our proposed mini-batching strategy outperforms the
ASGD with a constant batch size (for example, applied in
\cite{dekel2012optimal,gimpel2010distributed}). Note that such
strategy of increasing the batch size in later stage of training has
been also suggested and used in recent works in training large neural
networks, e.g.,~\cite{Goyal2017, Keskar2017}.

\section{Conclusion}
In this paper, we have developed stochastic modified equations (SMEs) to model the asynchronous 
stochastic gradient descent (ASGD) algorithms in the continuous-time limit. For quadratic loss 
functions, the resulting SME can be put into a Langevin equation with a solution known to converge 
to the unique invariant measure with a convergence rate dictated by the corresponding temperature. 
We utilize such information to compare with the momentum SGD and prove the ``asynchrony begets 
momentum'' phenomenon. For the general case, though the resulting SME does not have an explicitly 
known invariant measure, it still provides rather precise trajectory predictions for the discrete 
ASGD dynamics. Moreover, with SME available, we are able to find optimal hyper-parameters for 
ASGD algorithms by performing a moment analysis and leveraging the optimal control theory.

\section*{Funding}
J.A. is partially supported by the Gene Golub Research Fellowship.
J.L. is supported by the National Science Foundation under award DMS-1454939, and 
L.Y. is partially supported by the U.S. Department of Energy, Office of Science, 
Office of Advanced Scientific Computing Research, Scientific Discovery through 
Advanced Computing (SciDAC) program and the National Science Foundation under 
award DMS-1818449.



\begin{thebibliography}{10}

\bibitem{agarwal2011distributed}
A.~Agarwal and J.C. Duchi.
\newblock Distributed delayed stochastic optimization.
\newblock In {\em Advances in Neural Information Processing Systems}, pages
  873--881, 2011.

\bibitem{bernstein1929}
Serge Bernstein.
\newblock Sur les fonctions absolument monotones.
\newblock {\em Acta Math.}, 52:1--66, 1929.

\bibitem{bottou2016optimization}
L.~Bottou, F.E. Curtis, and J.~Nocedal.
\newblock Optimization methods for large-scale machine learning, 2016.
\newblock preprint, arXiv:1606.04838.

\bibitem{dekel2012optimal}
O.~Dekel, R.~Gilad-Bachrach, O.~Shamir, and L.~Xiao.
\newblock Optimal distributed online prediction using mini-batches.
\newblock {\em Journal of Machine Learning Research}, 13(Jan):165--202, 2012.

\bibitem{duchi2011adaptive}
J.C. Duchi, E.~Hazan, and Y.~Singer.
\newblock Adaptive subgradient methods for online learning and stochastic
  optimization.
\newblock {\em Journal of Machine Learning Research}, 12(Jul):2121--2159, 2011.

\bibitem{duchi2013estimation}
J.C. Duchi, M.I. Jordan, and B.~McMahan.
\newblock Estimation, optimization, and parallelism when data is sparse.
\newblock In {\em Advances in Neural Information Processing Systems}, pages
  2832--2840, 2013.

\bibitem{gimpel2010distributed}
K.~Gimpel, D.~Das, and N.A. Smith.
\newblock Distributed asynchronous online learning for natural language
  processing.
\newblock In {\em Proceedings of the Fourteenth Conference on Computational
  Natural Language Learning}, pages 213--222. Association for Computational
  Linguistics, 2010.

\bibitem{Goyal2017}
P.~Goyal, P.~Doll{\'a}r, R.~Girshick, P.~Noordhuis, L.~Wesolowski, A.~Kyrola,
  A.~Tulloch, Y.~Jia, and K.~He.
\newblock Accurate, large minibatch {SGD}: {T}raining {ImageNet} in 1 hour,
  2017.
\newblock arXiv preprint, arXiv:1706.02677.

\bibitem{hardt2015train}
M.~Hardt, B.~Recht, and Y.~Singer.
\newblock Train faster, generalize better: Stability of stochastic gradient
  descent.
\newblock In {\em International Conference on Machine Learning}, pages
  1225--1234, 2016.

\bibitem{Keskar2017}
N.S. Keskar, D.~Mudigere, J.~Nocedal, M.~Smelyanskiy, and P.~T.~P. Tang.
\newblock On large-batch training for deep learning: {G}eneralization gap and
  sharp minima.
\newblock In {\em International Conference on Learning Representations}, 2017.

\bibitem{kingma2014adam}
D.P. Kingma and J.~Ba.
\newblock Adam: A method for stochastic optimization.
\newblock {\em International Conference on Learning Representations}, 2015.

\bibitem{kloeden1992stochastic}
P.E. Kloeden and E.~Platen.
\newblock Stochastic differential equations.
\newblock In {\em Numerical Solution of Stochastic Differential Equations},
  pages 103--160. Springer, 1992.

\bibitem{li2015stochastic}
Q.~Li, C.~Tai, and W.~E.
\newblock Stochastic modified equations and adaptive stochastic gradient
  algorithms.
\newblock In {\em International Conference on Machine Learning}, pages
  2101--2110, 2017.

\bibitem{liu2015asynchronous2}
J.~Liu and S.J. Wright.
\newblock Asynchronous stochastic coordinate descent: Parallelism and
  convergence properties.
\newblock {\em SIAM Journal on Optimization}, 25(1):351--376, 2015.

\bibitem{liu2015asynchronous}
J.~Liu, S.J. Wright, C.~R{\'e}, V.~Bittorf, and S.~Sridhar.
\newblock An asynchronous parallel stochastic coordinate descent algorithm.
\newblock {\em The Journal of Machine Learning Research}, 16(1):285--322, 2015.

\bibitem{mitliagkas2016asynchrony}
I.~Mitliagkas, C.~Zhang, S.~Hadjis, and C.~R{\'e}.
\newblock Asynchrony begets momentum, with an application to deep learning.
\newblock In {\em Communication, Control, and Computing (Allerton), 2016 54th
  Annual Allerton Conference on}, pages 997--1004. IEEE, 2016.

\bibitem{needell2014stochastic}
D.~Needell, R.~Ward, and N.~Srebro.
\newblock Stochastic gradient descent, weighted sampling, and the randomized
  kaczmarz algorithm.
\newblock In {\em Advances in Neural Information Processing Systems}, pages
  1017--1025, 2014.

\bibitem{nesterov2012efficiency}
Y.~Nesterov.
\newblock Efficiency of coordinate descent methods on huge-scale optimization
  problems.
\newblock {\em SIAM Journal on Optimization}, 22(2):341--362, 2012.

\bibitem{pavliotis2014stochastic}
G.A. Pavliotis.
\newblock {\em Stochastic processes and applications: Diffusion processes, the
  Fokker-Planck and Langevin equations}, volume~60.
\newblock Springer, 2014.

\bibitem{polyak1964some}
B.T. Polyak.
\newblock Some methods of speeding up the convergence of iteration methods.
\newblock {\em USSR Computational Mathematics and Mathematical Physics},
  4(5):1--17, 1964.

\bibitem{recht2011hogwild}
B.~Recht, C.~R{\'e}, S.~Wright, and F.~Niu.
\newblock Hogwild: A lock-free approach to parallelizing stochastic gradient
  descent.
\newblock In {\em Advances in neural information processing systems}, pages
  693--701, 2011.

\bibitem{richtarik2014iteration}
P.~Richt{\'a}rik and M.~Tak{\'a}{\v{c}}.
\newblock Iteration complexity of randomized block-coordinate descent methods
  for minimizing a composite function.
\newblock {\em Mathematical Programming}, 144(1-2):1--38, 2014.

\bibitem{sutskever2013importance}
I.~Sutskever, J.~Martens, G.~Dahl, and G.~Hinton.
\newblock On the importance of initialization and momentum in deep learning.
\newblock In {\em International conference on machine learning}, pages
  1139--1147, 2013.

\bibitem{tieleman2012lecture}
T.~Tieleman and G.~Hinton.
\newblock Lecture 6.5-rmsprop: Divide the gradient by a running average of its
  recent magnitude.
\newblock {\em COURSERA: Neural networks for machine learning}, 4(2):26--31,
  2012.

\bibitem{younes2005verification}
HL~Younes.
\newblock {\em Verification and Planning for Stochastic Processes with
  Asynchronous Events}.
\newblock PhD thesis, Academy of Engineering Sciences, 2005.

\end{thebibliography}

\section*{Appendix A: miscellaneous computations in SMEs}
In this section, we provide the missing computations in Section 2.

\subsection{Evolution equation of $\Sigma$ for nonlinear gradients}
First, we have
\begin{align*} \Sigma_k &= 
  \mathbb{E}\bigg((-\frac{y_k}{\alpha} - \nabla f_{\gamma_k}(x_{k-\tau_k}))(-\frac{y_k}{\alpha} - \nabla
  f_{\gamma_k}(x_{k-\tau_k}))^T\bigg).
\end{align*}
By expanding the terms in the expectation and treating them individually, we arrive at the following
\begin{equation}\label{eq:sigma}
\begin{aligned}
  \Sigma_k &= \frac{1}{\alpha^2}y_k y_k^T+ \frac{y_k}{\alpha}\mathbb{E}\{ \nabla f_{\gamma_k} (x_{k-\tau_k})^T\} +\mathbb{E}\{ \nabla f_{\gamma_k} (x_{k-\tau_k})\}\frac{y_k^T}{\alpha} +\mathbb{E}\{\nabla f_{\gamma_k} (v_k)\nabla f_{\gamma_k} (x_{k-\tau_k})^T\} \\
  & = \mathbb{E}\{\nabla f_{\gamma_k} (x_{k-\tau_k})\nabla f_{\gamma_k} (x_{k-\tau_k})^T\} - \frac{1}{\alpha^2}y_k y_k^T\\
  & = \mu \sum_{m=0}^{\infty}\mathbb{E}\{\nabla f_{\gamma_{k-1}} (x_{k-1-m})\nabla f_{\gamma_{k-1}} (x_{k-1-m})^T\}(1-\mu)\mu^m\\
  & \,\,\,\, +(1-\mu) \mathbb{E}\{\nabla f_{\gamma_k}(x_k)\nabla f_{\gamma_k}(x_k)^T \}- \frac{1}{\alpha^2}y_k y_k^T\\
  & = \mu \big(\Sigma_{k-1} +\frac{1}{\alpha^2}y_{k-1} y_{k-1}^T\big) +(1-\mu) \mathbb{E}\{\nabla f_{\gamma_k}(x_k)\nabla f_{\gamma_k}(x_k)^T \}- \frac{1}{\alpha^2}y_k y_k^T\\
  & = \mu \big(\Sigma_{k-1} +\frac{1}{\alpha^2}y_{k-1} y_{k-1}^T\big) +\frac{1-\mu}{n}\sum_{i=1}^n\nabla f_i(x_k)\nabla f_i(x_k)^T - \frac{1}{\alpha^2}y_k y_k^T.
\end{aligned}
\end{equation}
Notice that $y_k = \mu y_{k-1} - \alpha(1-\mu) \nabla f(x_k)$, and thus we have
\begin{align*}
  y_{k-1}y_{k-1}^T &= \frac{1}{\mu^2} \big( y_k +\alpha(1-\mu) \nabla f(x_k)\big)\big( y_k
  +\alpha(1-\mu) \nabla f(x_k)\big)^T\\ &= \frac{1}{\mu^2} \bigg( y_k y_k^T + \alpha(1-\mu) y_k \nabla
  f(x_k)^T + \alpha(1-\mu) \nabla f(x_k) y_k^T + \alpha^2(1-\mu)^2 \nabla f(x_k)\nabla f(x_k)^T \bigg).
\end{align*}
Substituting it in \eqref{eq:sigma}, we obtain
\begin{align*}
  \frac{\Sigma_k - \Sigma_{k-1}}{\alpha(1-\mu)} =& -\frac{1}{\alpha}\Sigma_{k-1} +\frac{1}{\alpha^3\mu} y_k y_k^T + \frac{1}{\alpha^2\mu} y_k \nabla f(x_k)^T + \frac{1}{\alpha^2\mu} \nabla f(x_k) y_k^T \\
  & + \frac{1-\mu}{\alpha\mu} \nabla f(x_k)\nabla f(x_k)^T + \frac{1}{\alpha n}\sum_{i=1}^n\nabla f_i(x_k)\nabla f_i(x_k)^T.
\end{align*}
Using this and $\Delta t = \alpha(1-\mu), \alpha = \sqrt{\frac{\eta}{1-\mu}}$, we obtain the
evolution equation \eqref{evol}.

\subsection{Evolution equation of $\Sigma$ for linear gradients}
Similar to section 5.1, we have
\begin{align}\label{sig}
\nonumber    \Sigma_k &= \mathbb{E}\bigl((\nabla f(m_k) - \nabla f_{\gamma_k}(x_{k-\tau_k}))(\nabla f(m_k) - \nabla
    f_{\gamma_k}(x_{k-\tau_k}))^T\bigr)\\
    &= \mu(\Sigma_{k-1}+\nabla f(m_{k-1})\nabla f(m_{k-1})^T) + (1-\mu) \mathbb{E}(\nabla f_{\gamma_k}(x_k)\nabla f_{\gamma_k}(x_k)^T)-\nabla f(m_k)\nabla f(m_k)^T.
\end{align}
We then subtract both sides by $\Sigma_{k-1}$ and divide by $\Delta t = \sqrt{\eta(1-\mu)}$. Moreover, we use the relation
\begin{align*}
    x_k = \frac{m_k-\mu m_{k-1}}{1-\mu} = \frac{m_k - \mu(m_k-p_k\Delta t)}{1-\mu} = m_k + \mu\sqrt{\frac{\eta}{1-\mu}} p_k
\end{align*}
to replace $x_k$ in (\ref{sig}), and replace $m_{k-1}$ by $m_k-p_k\Delta t$. Then, since the gradient of $f$ is linear, rearrange the terms we get 
\begin{align*}
    \frac{\Sigma_k - \Sigma_{k-1}}{\sqrt{\eta(1-\mu)}} =& -\sqrt{\frac{1-\mu}{\eta}} \Sigma_{k-1} - \sqrt{\frac{1-\mu}{\eta}}\nabla f(m_k)\nabla f(m_k)^T + \mu \sqrt{\eta(1-\mu)} \nabla f(p_k)\nabla f(p_k)^T\\
    -\mu(\nabla f(m_k)\nabla f(p_k)^T &+\nabla f(p_k)\nabla f(m_k)^T) + \frac{1}{n}\sqrt{\frac{1-\mu}{\eta}}\sum_{i=1}^{n}\nabla f_i(m_k + \mu\sqrt{\frac{\eta}{1-\mu}} p_k)\nabla f_i(m_k + \mu\sqrt{\frac{\eta}{1-\mu}} p_k)^T.
\end{align*}

\subsection{SME for SGD with momentum}
Recall the iteration for the SGD with a constant momentum parameter is
\begin{align*}
  &v_{k+1} = \mu' v_k - \eta' \nabla f_{\gamma_k}(x_k) \nonumber\\
  & x_{k+1} = x_k +  v_{k+1},
\end{align*}
which can be viewed as a second-order difference equation. To ensure the final equation with all
terms of order $O(1)$, one needs $\eta' = (\Delta t)^2$. We can rewrite \eqref{momentum} as
\begin{align}
  &\frac{v_{k+1}}{\sqrt{\eta'}} = \frac{v_k}{\sqrt{\eta'}} +\sqrt{\eta'} \bigg( -\frac{1-\mu'}{\eta'} v_k - \nabla f(x_k) \bigg) + \sqrt{\eta'}(\nabla f(x_k)- \nabla f_{\gamma_k}(x_k))\nonumber\\
  & x_{k+1} = x_k +  \frac{v_{k+1}}{\sqrt{\eta'}} \sqrt{\eta'}.
\end{align}
Let us introduce $p = v/\sqrt{\eta'}$. In order to have $\sqrt{\eta'} (\nabla
f(x_k)-\nabla_{\gamma_k} f(x_k) )\sim c \Delta B_t$, we choose $c \sim \sigma
(\eta')^{1/4}$. Therefore, we obtain the first order weak approximation, which can also be viewed as
the Euler-Maruyama discretization of the following SDE
\begin{align*}
  & dP_t = -\nabla f(X_t) dt - \frac{1-\mu'}{\sqrt{\eta'}}P_t dt + \sigma(X_t)(\eta')^{\frac{1}{4}}
  dB_t \nonumber \\ & dX_t = P_t dt.
\end{align*}

\section*{Appendix B: dynamics of SME-ASGD \eqref{nonlinearASGD}}

We consider the one dimensional case with $f(x) = \frac{1}{2}a x^2$.
The goal here is to give an analysis of the dynamics of first and
second moment of $X$ and $Y$ under \eqref{nonlinearASGD}. Taking
expectation, we obtain
\begin{align*}
  d \begin{bmatrix} \mathbb{E}(Y_t)\\ \mathbb{E}(X_t) \end{bmatrix} =
  \begin{bmatrix} -\sqrt{\frac{1-\mu}{\eta}} & -a \\ 1 & 0 \end{bmatrix}
  \begin{bmatrix} \mathbb{E}(Y_t)\\    \mathbb{E}(X_t) \end{bmatrix}
  dt  =
  A(\mu, \eta)\begin{bmatrix} \mathbb{E}(Y_t)\\ \mathbb{E}(X_t) \end{bmatrix} dt.
\end{align*}
One observes that the eigenvalues of $A(\mu, \eta)$ are
$\lambda_{1,2}(A) = \frac{1}{2} \bigg(-\sqrt{\frac{1-\mu}{\eta} }\pm
\sqrt{\frac{1-\mu}{\eta}-4a} \bigg)$,
the real parts of both are negative as long as $a>0$. From this, we
conclude that, when $a>0$, the expectation of $X_t$ decays
exponentially. The corresponding stationary solutions are given by
\[
\mathbb{E}(X_{\infty}) = \mathbb{E}(Y_{\infty}) = 0.
\]
For the second moment, we end up with the following equations by using the Ito's formula
\begin{align}\label{spec2}
  \nonumber
  &d\mathbb{E}(X_t^2) = 2\mathbb{E}(X_tY_t) dt +\Sigma(t) \frac{\eta^{3/2}}{(1-\mu)^{1/2}} dt\\
  \nonumber
  &d\mathbb{E}(Y_t^2) = -2a\mathbb{E}(X_tY_t) dt -2\sqrt{\frac{1-\mu}{\eta}}\mathbb{E}(Y_t^2) dt \\
  &d\mathbb{E}(X_t Y_t)=-a\mathbb{E}(X_t^2)dt + \mathbb{E}(Y_t^2) dt  -  \sqrt{\frac{1-\mu}{\eta}} \mathbb{E}(X_tY_t) dt.
\end{align}
In order to study the behavior of the second moments, we can rewrite \eqref{spec2} as
\begin{align}\label{moment}
  d\begin{bmatrix} \mathbb{E}(X_t^2)\\ \mathbb{E}(Y_t^2) \\ \mathbb{E}(X_t Y_t) \end{bmatrix} &= \begin{bmatrix} 0 & 0 & 2\\ 0 & -2\sqrt{\frac{1-\mu}{\eta}} & -2a \\ -a & 1 & -\sqrt{\frac{1-\mu}{\eta}}  \end{bmatrix}\begin{bmatrix} \mathbb{E}(X_t^2)\\ \mathbb{E}(Y_t^2) \\ \mathbb{E}(X_t Y_t) \end{bmatrix} dt  + \begin{bmatrix} \Sigma(t) \frac{\eta^{3/2}}{(1-\mu)^{1/2}} \\ 0 \\ 0\end{bmatrix}  dt.
\end{align}
The corresponding stationary solutions are
\[
\mathbb{E}(X_{\infty}Y_{\infty}) = \frac{-\Sigma \eta^{3/2}}{2(1-\mu)^{1/2}}, \,\, \mathbb{E}(Y_{\infty}^2) = \frac{a\Sigma \eta^2}{2(1-\mu)} , \, \text{ and } \mathbb{E}(X_{\infty}^2) =\frac{\Sigma \eta^2}{2(1-\mu)}+\frac{\Sigma\eta}{2 a}.
\]

Let us introduce
\[
B(\mu, \eta) = \begin{bmatrix} 0 & 0 & 2\\ 0 & -2\sqrt{\frac{1-\mu}{\eta}} & -2a \\ -a & 1 &
  -\sqrt{\frac{1-\mu}{\eta}} \end{bmatrix}.
\]
The eigenvalues of $B(\mu, \eta)$ are
\[
\lambda_1 = -\sqrt{\frac{1-\mu}{\eta}}, \lambda_{2,3} = \lambda_{\pm} = -\sqrt{\frac{1-\mu}{\eta}}
\pm \sqrt{\frac{1-\mu}{\eta}-4a}.
\]
We can see that the real parts of all roots are negative as long as
$a>0$. Moreover, the second moment of $X_t$ decays
exponentially, with the rate given by $\mathrm{Re}(\lambda_+)$ since
$\lambda_+$ is the eigenvalue with the largest (negative) real
part. We obtain the largest descent rate $\mathrm{Re}(\lambda_+)$ when
the second part $\sqrt{\frac{1-\mu}{\eta}-4a}$ in $\lambda_+$ is
purely imaginary, i.e., when $\mu $ takes
\begin{align}\label{optmu}
  \mu_{\text{opt}} = \max\{1-4a\eta, 0\}.
\end{align}
We note that \eqref{optmu} also gives a suggestion to choose optimal
step size $\eta$: when $\mu$ is given, the maximal step size we can
choose is $\eta _{\text{opt}}= \frac{1-\mu}{4a}$. Any step size beyond
that will cause oscillations in the SME and the corresponding ASGD.

\end{document}